\documentclass{article}

\PassOptionsToPackage{numbers, compress}{natbib}

\usepackage[preprint]{neurips_2020}

\usepackage[utf8]{inputenc} 
\usepackage[T1]{fontenc}    
\usepackage{hyperref}       
\usepackage{url}            
\usepackage{booktabs}       
\usepackage{amsfonts}       
\usepackage{amsmath}
\usepackage{nicefrac}       
\usepackage{microtype}      
\usepackage{graphicx}
\usepackage{subcaption}
\usepackage[ruled]{algorithm2e}

\usepackage{wrapfig}
\usepackage{color}

\newboolean{include-notes}
\setboolean{include-notes}{true}
\newcommand{\lcnote}[1]{\ifthenelse{\boolean{include-notes}} {{\color{green}LC:\@ #1}}{}}
\newcommand{\hgnote}[1]{\ifthenelse{\boolean{include-notes}} {{\color{blue}HG:\@ #1}}{}}

\newcommand{\demonstrator}{demonstrator}
\newcommand{\inferrer}{inferrer}

\DeclareMathOperator*{\argmax}{argmax}

\graphicspath{ {./img/} }

\title{Accounting for Human Learning when Inferring Human Preferences}

\author{
  Harry Giles\\
  Centre for Human Compatible AI\\
  \texttt{harry@humancompatible.ai} \\
  \And{}
  Lawrence Chan \\
  UC Berkeley \\
  \texttt{chanlaw@berkeley.edu} \\
}

\begin{document}

\maketitle

\begin{abstract}
Inverse reinforcement learning (IRL) is a common technique for inferring human preferences from data. Standard IRL techniques tend to assume that the human demonstrator is stationary, that is that their policy $\pi$ doesn't change over time. In practice, humans interacting with a novel environment or performing well on a novel task will change their demonstrations as they learn more about the environment or task. We investigate the consequences of relaxing this assumption of stationarity, in particular by modelling the human as learning. Surprisingly, we find in some small examples that this can lead to better inference than if the human was stationary. That is, by observing a demonstrator who is themselves learning, a machine can infer \emph{more} than by observing a demonstrator who is noisily rational. In addition, we find evidence that misspecification can lead to poor inference, suggesting that modelling human learning is important, especially when the human is facing an unfamiliar environment.
\end{abstract}

\section{Introduction}

Directly specifying the desired task that a machine should perform can be challenging~\cite{lehman2020surprising}. One approach to this problem is through Inverse Reinforcement Learning (IRL), where the task is to learn a reward function $R$ which best explains the behaviour of a given (noisily-)optimal agent $\pi$~\cite{ng00-algor-inver-reinf-learn}, of which the standard model is the Boltzmann-rational model~\cite{ziebart08-maxim-entrop-inver-reinf-learn, ziebart2010modeling, finn2016guided,  christiano2017deep, fu2017learning, ibarz2018reward,brown2019extrapolating}.



One strong assumption that we challenge in this work is that of \textit{stationarity}. Models of near-optimality tend to assume that the agent's demonstrations are stationary; that is, that the policy does not change over time. This may not be the case for several reasons, as the following examples show:
\begin{enumerate}
\item For a robotics task, the human demonstrator's performance might improve during demonstrations, as they become quicker to complete a task as their understanding of the dynamics develops, or as they explore and then exploit better strategies
\item With recommender systems, a human might still be learning their preferences; they might not know how much they enjoy a certain genre of music until they have tried it.
\item When labelling images~\cite{deng2009imagenet} or ranking demonstrations~\cite{christiano2017deep, brown2019extrapolating}, a human labeller might not understand the instructions precisely at first, but then they develop a better understanding by interacting with more examples.
\end{enumerate}
In each of these examples, while a true human \emph{expert} might not change their behavior through the course of generating demonstrations or comparisons, a human \demonstrator{} interacting with a novel system might learn through the course of interacting.

Our key proposal is to account for non-stationary by modelling the demonstrator as learning. We first investigate how this modelling choice interacts with IRL\@. Is there anything a machine \inferrer{} can learn from human demonstrators that are themselves learning? Our results show that the answer is yes. Moreover, surprisingly we find that non-stationary learning demonstrators can be more informative across several domains. Further, we shed light on this result by showing that a learning \demonstrator{} is able to disambiguate between rewards that stationary demonstrators cannot. Next, we investigate the impacts of model misspecification in relation to stationarity, and find evidence that assuming stationarity is harmful when the human \demonstrator{} is learning. Encouragingly, we find that merely modelling the \demonstrator{} as learning is sufficient for the \inferrer{} to get good performance in most cases, even if the \inferrer{} gets the parameters of the learning algorithm incorrect.

\subsection{Related work}
{A large amount of recent work studies IRL with alternative assumptions. To list a few, \citet{bobu2020less} uses a modified version of the Boltzmann-rationality assumption; \citet{evans2016learning} presents an algorithm for learning from various psychology-inspired algorithms; \citet{reddy2018you} studies the problem of IRL from an agent with false dynamics beliefs; \citet{zhi2020online} presents an algorithm from learning from the trajectories of bounded planning agents; and \citet{levine2012continuous} studies IRL under the assumption that the demonstrations are merely \emph{locally} (noisily-){optimal}.} However, in the vast majority of work in this area, the agents are assumed to be stationary and don't change their revealed preferences over time.

\citet{chan2019assistive} and \citet{jacq2019learning} do study versions of the task of learning from a learner. However, \citeauthor{chan2019assistive}'s Assistive Bandits studies the problem in the restricted domain of a multi-armed bandit, and thus makes the assumption that the human only learns about the reward and not the dynamics. \citeauthor{jacq2019learning}'s Learning from a learner presents an algorithm that relies on the assumption that later actions have higher Q-value than earlier actions and then performs IRL from comparisons on top of that. This work both considers sequential decision problems where the demonstrator does not know the dynamics and makes a different assumption about the learning algorithms.
\section{Background}\label{sec:back}
A finite \textit{Markov Decision Process} is a tuple $\mathcal{M} = (\mathcal{S}, \mathcal{A}, T, \rho, \gamma, R_\theta, P_0)$ with
\begin{itemize}
\item Finite sets of states $\mathcal{S}$ and actions $\mathcal{A}$
\item Transition probabilities $T(s, a, s')$ over $s' \in \mathcal{S}$, $\forall s \in \mathcal{S}, a \in \mathcal{A}$
\item Initial state probabilities $\rho$ over $\mathcal{S}$
\item Discount rate $\gamma \in [0, 1)$
\item Parameterised reward function $R_\theta : \mathcal{S} \times \mathcal{A} \rightarrow \mathbb{R}$, for parameter $\theta \in \Theta$
\item Prior distribution over $\theta$, $P_0$
\end{itemize}
A \textit{policy} $\pi(a \mid s)$ gives a probability vector over actions $a \in \mathcal{A}$ for a given state $s\in \mathcal{S}$. For policy $\pi$, the action-value function $Q^\pi : \mathcal{S} \times \mathcal{A} \rightarrow \mathbb{R}$ is the expected return of starting in state $s$, taking action $a$, and then following policy $\pi$:

\[Q^\pi(s, a) = \mathbb{E}\left[ R(s, a) + \gamma R(s', a') + \ldots | \pi, T\right]\]

The optimal action-value function $Q^*$ is defined to be $Q^*(s, a) = \max_\pi Q^\pi(s, a)$, and the optimal policy is $\pi^*(\cdot \mid s) \sim \text{Unif}\{ \argmax_a Q^*(s, a) \}$.

A \textit{planner} is a map $D : \Theta \rightarrow \Pi$ from reward parameter $\theta$ to policy $\pi$. The type of planner is assumed before doing IRL\@. The optimal planner $D^*$ has $D^*(\theta) = \pi^*$, the optimal policy. The Boltzmann planner has $D^\beta(\theta) = \pi^\beta$, where $\pi^\beta$ is the Boltzmann policy parameterised by $\beta$, defined to be the policy satisfying\footnote{This does not necessarily define a unique policy, see for example \citet{asadi16-alter-softm-operat-reinf-learn}.}

\[\pi(a \mid s) = \frac{\exp \beta Q^\pi(s, a)}{\sum_{a'} \exp \beta Q^\pi(s, a') }\]

\begin{wrapfigure}{r}{0.45\textwidth}
  \centering
  \includegraphics[width = 0.4\textwidth]{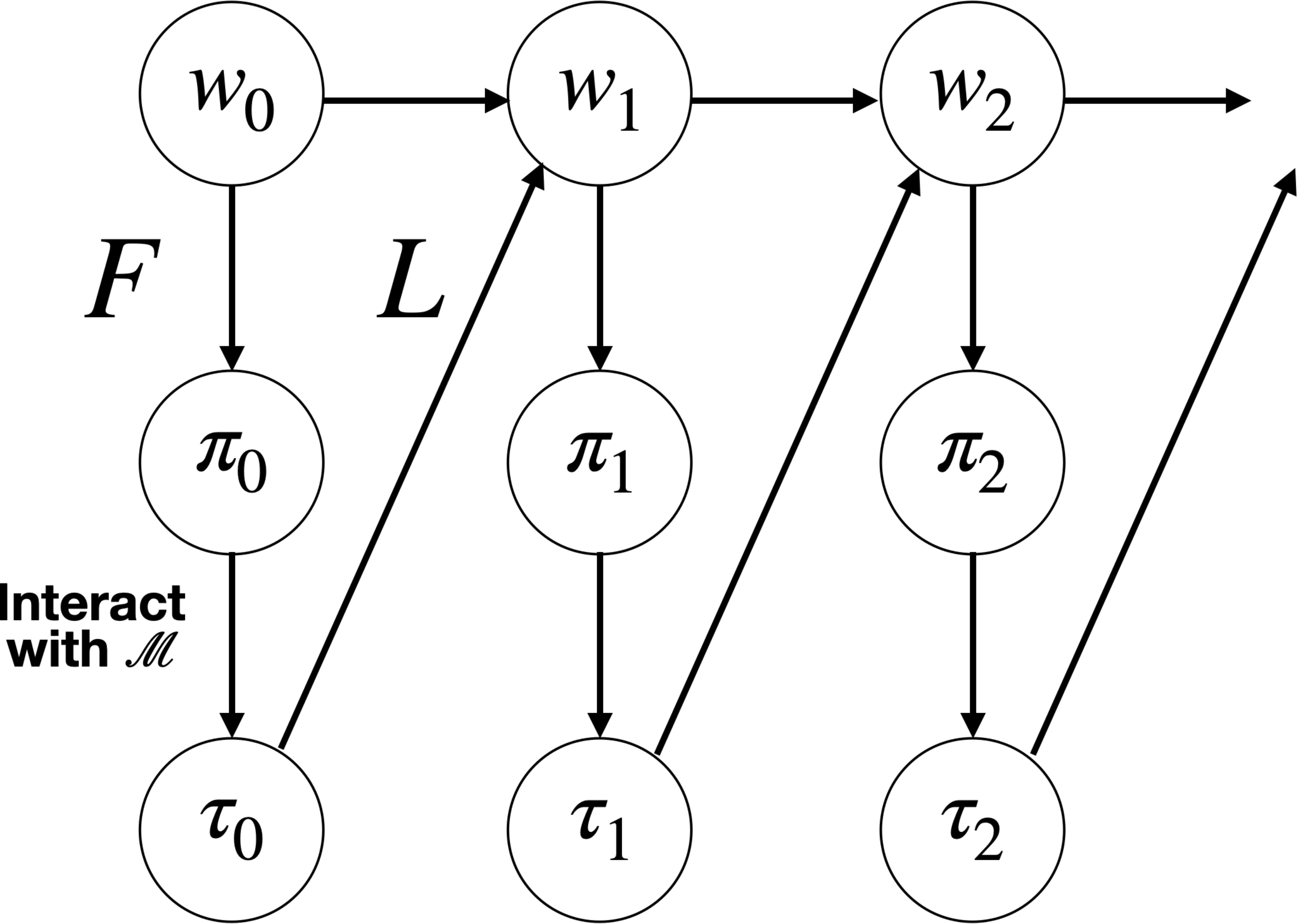}
  \caption{Diagram of a learning algorithm, as discussed in Section~\ref{sec:back}}\label{fig:learners-diagram}
\end{wrapfigure}

The form of the IRL problem which we shall address is not that which has access to an entire policy $\pi$, but that which has access to only to a series of $n$ demonstrations $(\tau_0, \ldots, \tau_n)$ which have been generated according to $\pi$.

As the primary form of non-stationarity we'd like to address is learning, we propose the following model of a learning agent. Given some parameter space $W$, a \textit{learner} consists of a policy-generating-function $F : W \rightarrow \Pi$ which takes parameters $w$ to policy $\pi \in \Pi$, paired with an update-function $L : W \times \{\tau\} \rightarrow W$ for performing parameter updates, where $\{\tau\}$ is the set of all single trajectories. The updates occur once at the end of each trajectory. A diagram of the learning process being assumed is given in Figure~\ref{fig:learners-diagram}. Note that our formulation technically allows for arbitrary changes in policies, however, in this work we restrict our attention to ``reasonable'' learning algorithms.

\section{Methods}

\begin{algorithm}[t]
\caption{BIRL variant for non-stationary agents}\label{alg:birl}
\KwIn{Demonstrations $(\tau_0, \ldots, \tau_n)$, prior $P_0(\theta)$, initial parameters $w_0$, learning algorithm $(F, L)$}

Initialise parameters $w_\theta = w_0$, $\mathbb{P}(\theta) = P_0(\theta)$

\For{$\tau$ in $(\tau_0, \ldots, \tau_n)$}{
    $\pi_\theta \leftarrow F(w_\theta)$\;
    Compute log-likelihood $\ell(\theta) \leftarrow \sum_{(s, a) \in \tau} \log \pi_\theta(a \mid s)$\;
    Update posterior $\log \mathbb{P}(\theta) \leftarrow \log \mathbb{P}(\theta) + \ell(\theta)$\;
    Fill in the reward according to each $\theta$, $\tau_\theta$\;
    Update agents $w_\theta \leftarrow L(w_\theta, \tau_\theta)$
}
\end{algorithm}

To perform inference on $\theta$, we use a variant on Bayesian Inverse Reinforcement Learning (BIRL)~\cite{ramachandran07-bayes-inver-reinf-learn}. Given a sequence of trajectories $ (\tau_0, \tau_2, \ldots, \tau_n)$, we compute
\[ \mathbb{P}(\theta|\tau_0, \tau_2, \ldots, \tau_n) = \frac{\mathbb{P}(\tau_0, \tau_2, \ldots, \tau_n|\theta) P_0(\theta)}{\sum_{\theta' \in \Theta} \mathbb{P}(\tau_0, \tau_2, \ldots, \tau_n | \theta')P_0(\theta')}\]


A general algorithm is detailed in Algorithm~\ref{alg:birl}.

\subsection{Exploring inference performance in simulation}\label{sec:why-simulation}
In an ideal world, we would be able to know a priori the reward parameters and learning algorithms that real human demonstrators are using, and then compare the performance of an \inferrer{} that accounts for the correct learning algorithm, models learning incorrectly, or fails to account for learning at all. However, because we don't have access to the true reward parameters and learning algorithms of any particular real human, we start by studying simulated demonstrators with known learning algorithms and policy-generating functions, on environments where the ground truth reward is known.

\subsection{Agents}\label{sec:agents}
For our stationary \demonstrator{}s we consider the Boltzmann planner $D^\beta$ from Section~\ref{sec:back}. Our non-stationary \demonstrator{}s use the parameters $w \in W$ to maintain a table of Q-value estimates, $\hat Q(s, a)$. These estimates are updated according to one of three different targets, corresponding with observed state $s_t$, action $a_t$, and reward $r_t$:
\begin{enumerate}
    \item \textbf{$L_q$ --- Q-learning} with target $y(s_t, a_t) = r_t + \gamma \hat V(s_{t+1})$
    \item \textbf{$L_d$ --- Direct Evaluation} with reward-to-go target $y(s_t, a_t) = r_t + \gamma r_{t+1} + \cdots + \gamma^{T - t} r_T$, for horizon $T$
    \item \textbf{$L_\lambda$ --- TD$(\lambda)$} with target $y(s_t, a_t) = r_t  + \gamma \left(\lambda y(s_{t+1}, a_{t+1}) + (1 - \lambda) \hat V(s_{t+1}) \right)$. This \demonstrator{} interpolates between the Q-learning and Direct Evaluation agents.
\end{enumerate}
Where for the current policy $\pi$, $\hat V(s) = \mathbb{E}[\hat Q(s, a) \mid a \sim \pi]$. Note that $L_q = L_{\lambda = 0}$ and $L_d = L_{\lambda = 1}$. For each state action pair visited in trajectory $\tau_i$, the parameters are updated towards the target $y(s, a)$ with a learning rate of $\frac{1}{N(s, a) + 1}$, where $N(s, a)$ is the number of visits to $(s, a)$:
\begin{equation}
    \hat Q(s, a) \leftarrow \frac{N(s, a)}{N(s, a)+1}\hat Q(s, a) + \frac{1}{N(s, a)+1} y(s, a)
\end{equation}

The policy function $F = F^\beta$ returns the Boltzmann policy corresponding with current Q-value estimates $\hat Q$. In this paper, we initialise our estimates to zero, $\hat Q_0 = 0$. In simple environments we expect this algorithm to converge to the Boltzmann policy $\pi^\beta$, but this is not guaranteed (for example because of insufficient exploration).

\subsection{Evaluation}\label{sec:metrics}
To evaluate the quality of inference we look at the mutual information between reward parameters $\theta$ and the trajectories $\tau$. This is a suitable evaluation metric as it captures the reduction of uncertainty in the reward parameter $\theta$ from having observed the demonstration:
\begin{equation*}
    M_i = H(\theta) - H(\theta \mid \tau_0, \ldots, \tau_{i - 1})
\end{equation*}

We also look at the mutual information of only one trajectory $m_i := H(\theta) - H(\theta \mid \tau_{i - 1})$, as it will be instructive later to see how the mutual information varies as the agent converges. In some cases we look at the posterior probability of the true theta $\theta^*$ per-timestep, $p_i = \mathbb{P}(\theta^* \mid \tau_{i-1})$, or cumulatively $P_i = \mathbb{P}(\theta^* \mid \tau_0, \ldots, \tau_{i-1})$. In both cases we consider the cumulative metric to be most important.

\section{Illustrative example}\label{sec:two-state-mdp}
\begin{figure}
    \centering
    \begin{subfigure}[m]{0.42\textwidth} 
        \centering
        \includegraphics[width=\textwidth]{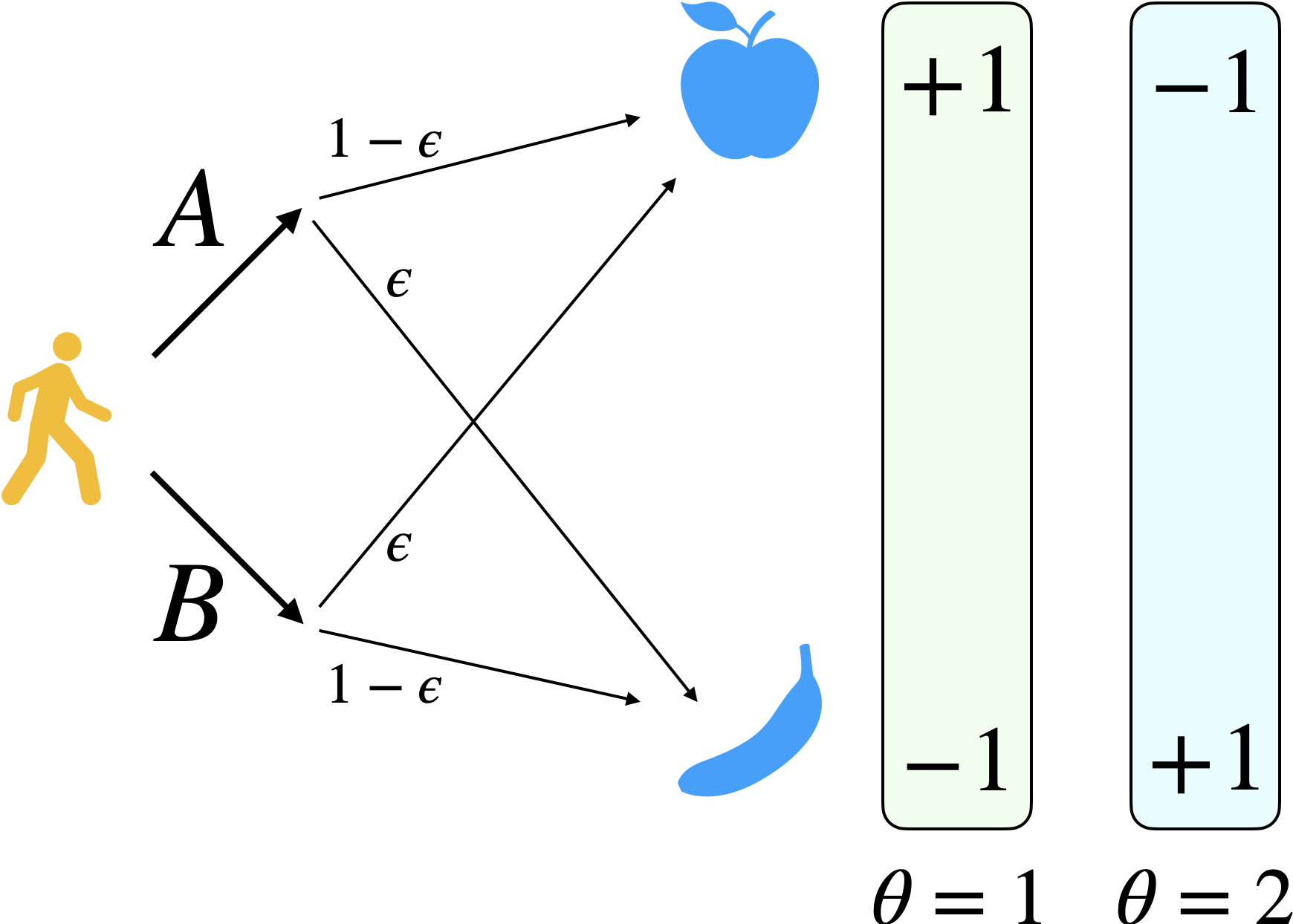}
    \end{subfigure}
    \begin{subfigure}[m]{0.55\textwidth}
        \centering
        \includegraphics[width=\textwidth]{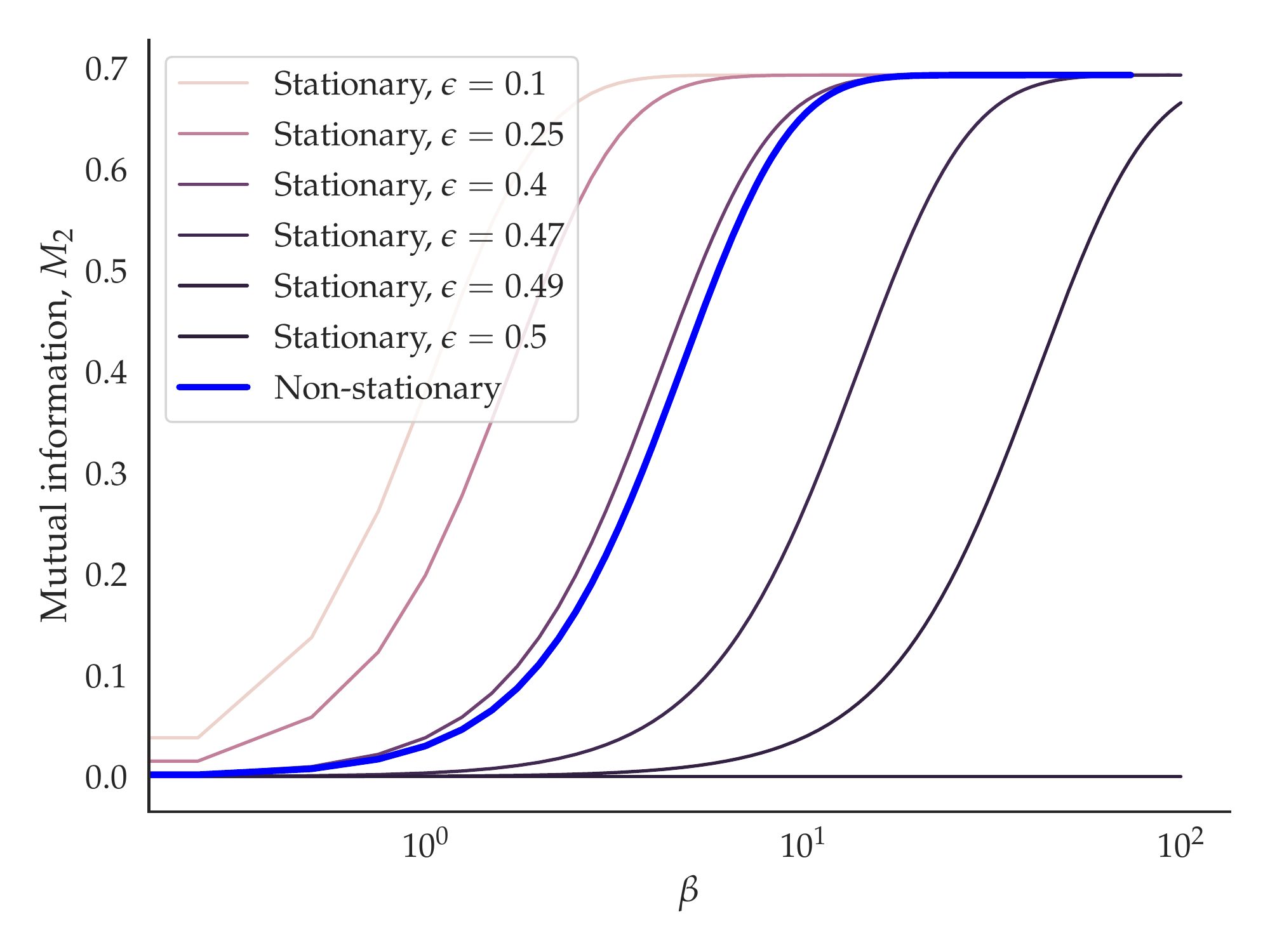}
    \end{subfigure}
    \caption{(Left) Diagram of the two-state MDP discussed in Section~\ref{sec:two-state-mdp}. (Right) The exact mutual information given by a demonstration of two trajectories as a function of the Boltzmann parameter $\beta$, plotted for a selection of noise parameters $\epsilon$. As discussed in Section~\ref{sec:two-state-mdp}, the mutual information of the stationary \demonstrator{} depends on the value of $\epsilon$, while the mutual information of the non-stationary learning \demonstrator{} is independent. In this case there exist values of $\epsilon$ such that the non-stationary demonstrator can be \emph{more informative} than the Boltzmann demonstrator.}\label{fig:two-state-mdp}
\end{figure}

Consider a simple two state MDP with two actions $A, B$ where with probability $1 - \epsilon$ $A$ leads to an apple and $B$ leads to a banana, and with probability $\epsilon$ the outcome is switched. Consider reward hypotheses $|\Theta| = 2$ where $\theta = 1$ specifies a preference for apples and $\theta = 2$ specifies a preference for bananas. See Figure~\ref{fig:two-state-mdp} for a diagram of this environment.

Over a demonstration of two trajectories, we compare a stationary agent --- the Boltzmann planner $D^\beta$ --- with a Direct Evaluation learner $(F^\beta, L_d)$. Note that the mutual information for a non-stationary agent in the first trajectory is always zero.

The mutual information is shown in Figure~\ref{fig:two-state-mdp} for varying temperature $\beta$ and noise $\epsilon$. Interestingly, the mutual information for the non-stationary agent does not depend on $\epsilon$: the agent will either update towards repeating the same action or update towards taking the opposite action, signalling their preference in each case. On the other hand, the performance of the stationary agent decreases for larger $\epsilon$; this is because the action has less influence on the expected reward. For small $\epsilon$ the stationary agent outperforms, and for large $\epsilon$ the non-stationary agent outperforms. In the case of $\epsilon = 0.5$, the action has no influence on the outcome, and therefore the stationary agent exhibits no mutual information. However, the non-stationary agent is still able to disambiguate in this case because they are signalling not only through their behaviour, but their change in behaviour.

\section{Experiments}
Motivated by our illustrative example, we compare the stationary and non-stationary models in three domains: an \textit{ambiguous} environment, adversarially designed such that the stationary policy is invariant among $\theta \in \Theta$, a Gridworld with more interpretable policies, and random MDPs.

In each experiment, we sample $\theta \sim \text{Unif}\{\Theta\}$, and record demonstrations for a stationary Boltzmann agent and a selection of the non-stationary agents described in Section~\ref{sec:agents}. Taking the demonstrations as input, we run BIRL, using the variant described in Algorithm~\ref{alg:birl} for the non-stationary agents. We repeat each experiment $n$ times and use a sample mean to estimate the metrics described in Section~\ref{sec:metrics}.

\subsection{Ambiguous MDP}\label{sec:ambiguous}
\begin{wrapfigure}{r}{0.5\textwidth}
    \centering
    \includegraphics[width = 0.4\textwidth]{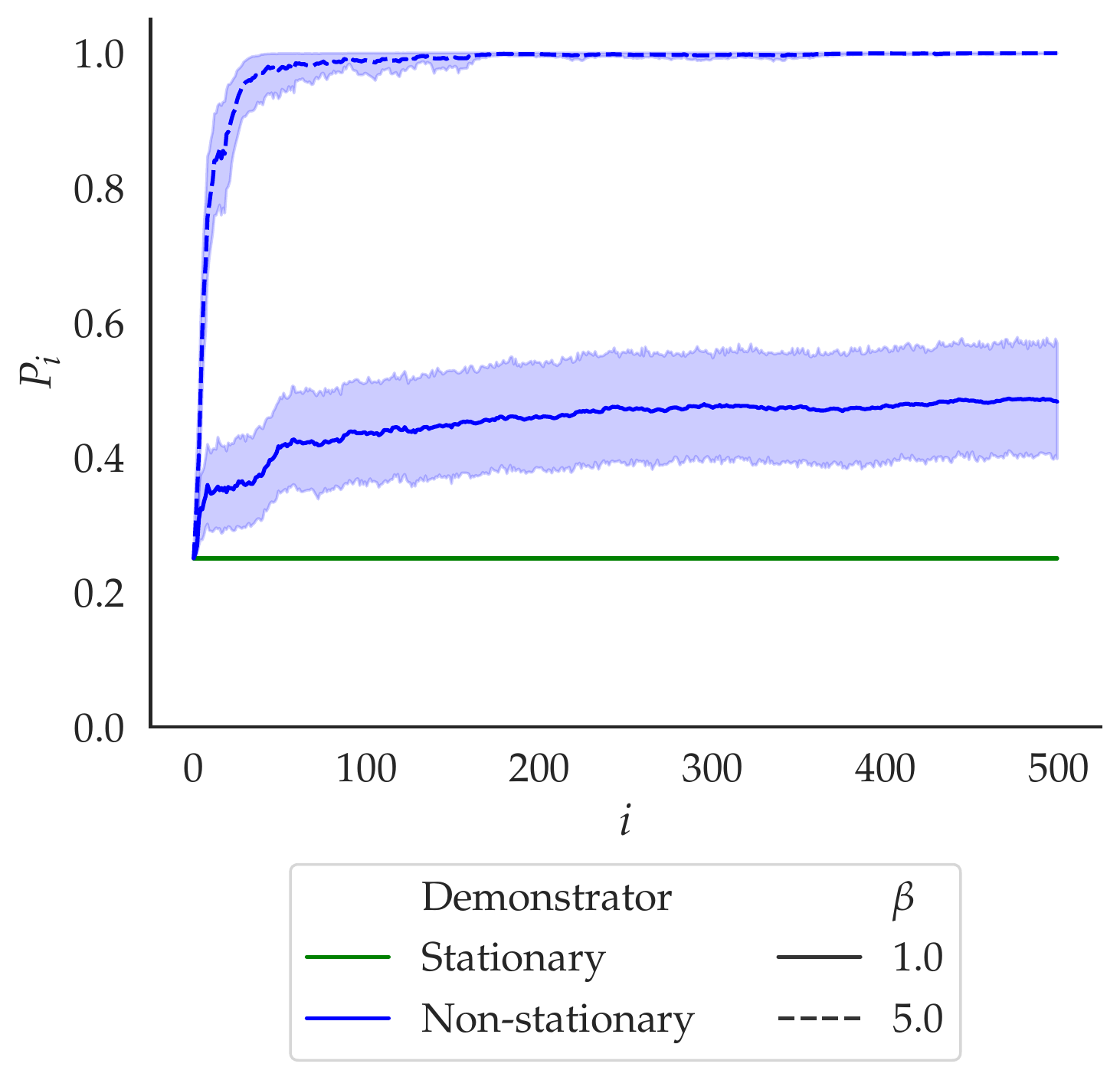}
    \caption{Ambiguous environment posterior probability $P_i$ (higher is better) for both a (non-stationary) Direct Evaluation \demonstrator{} compared with a stationary \demonstrator{}, for $\beta = 1, 5$, including $95\%$ bootstrapped confidence intervals. As discussed in Section~\ref{sec:ambiguous}, due to shaping, observing the stationary Boltzmann \demonstrator{} gives no information about the reward parameter.}\label{fig:ambiguous}
\end{wrapfigure}

In our illustrative example we saw that that learning demonstrators can be more informative, including the particular \textit{ambiguous} case where the stationary demonstrations don't depend on $\theta$. We further verify this observation here through use of reward shaping~\cite{ng99-polic-invar-under-rewar-trans}. We consider an MDP $|\mathcal{S}| = 4, |\mathcal{A}| = 3$ with arbitrarily chosen transition dynamics. We consider an arbitrary reward function which we shape three times, $|\Theta| = 4$, such that the Q-function in each state $Q(s, \cdot)$ is invariant up to a constant factor. In particular this means that the stationary Boltzmann planner does not depend on the rewar parameter $\theta$.

We run $n = 10$ experiments, each with 500 trajectories, and give the posterior at the true reward parameter $\theta^*$ in Figure~\ref{fig:ambiguous}. We observe that the non-stationary agent exhibits enough information for good reward inference, converging to a posterior of 1 in the case of $\beta = 5$, whereas in each case the stationary agent exhibits none. In conclusion we see again that through changing behaviour the learner demonstrator is able to signal it's preferences where the stationary demonstrator cannot.

\subsection{Gridworld experiment}\label{sec:gridworld}

\begin{wrapfigure}{r}{0.4\textwidth}
    \centering
    \includegraphics[width = 0.3\textwidth]{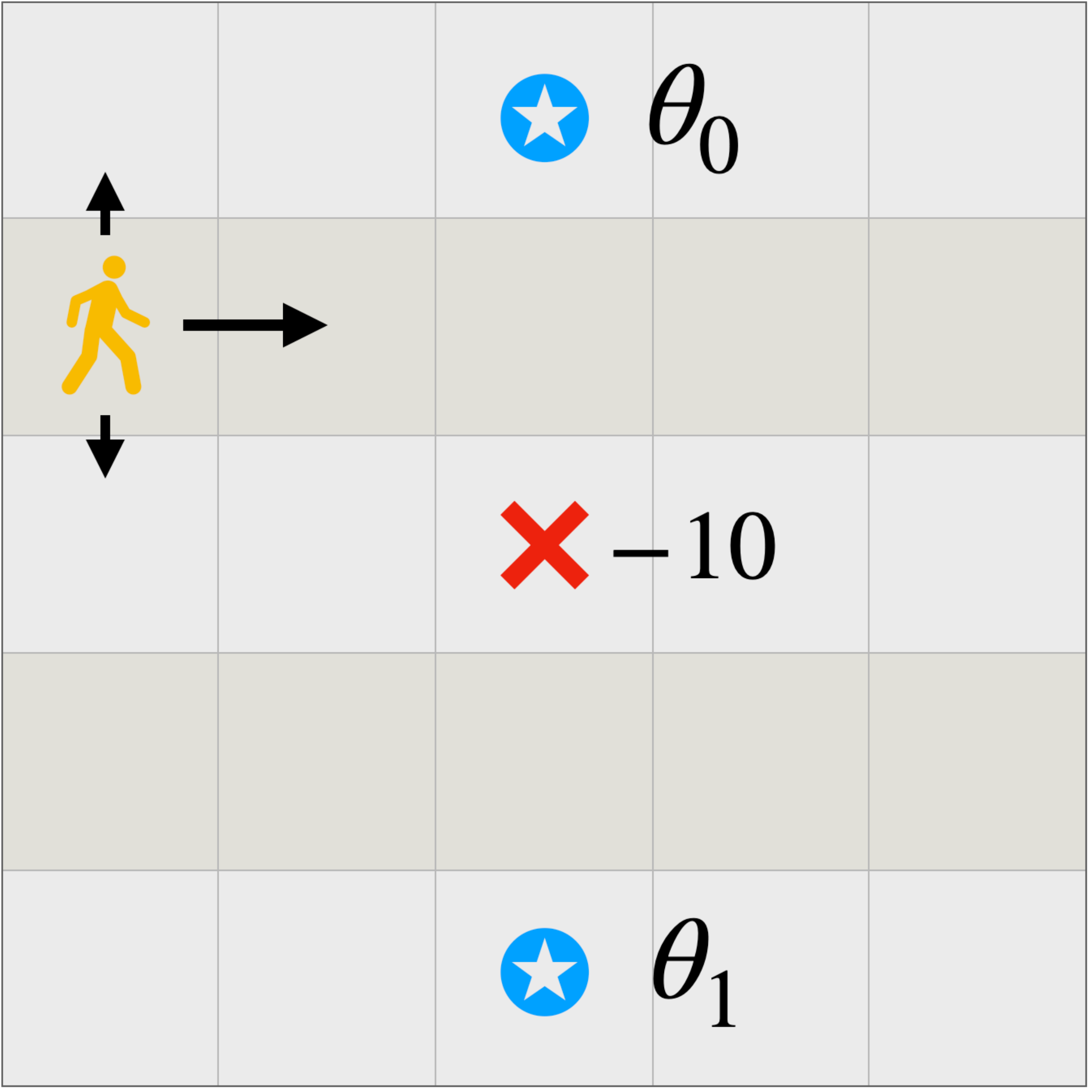}
    \caption{A visualisation of our Gridworld environment discussed in Section~\ref{sec:gridworld}.}\label{fig:gridworld}
\end{wrapfigure}

Next we turn to study the implications from our illustrative example in a more natural setting. We consider a $5 \times 5$ Gridworld environment based on OpenAI Gym's `Frozen-Lake-v0'~\cite{brockman16-openai-gym}. The environment has one hole state and two reward states. The reward parameter $\theta = (\theta_1, \theta_2)$ has $\theta_j \in \{0, 1, 2, 3, 4\}$, giving $|\Theta| = 25$. A parameter $\epsilon$ controls the noise in the environment: with probability $\epsilon$ the direction taken is not the direction chosen by the agent, but one of the two perpendicular directions. The environment is depicted in Figure~\ref{fig:gridworld}.

We run $n = 150$ experiments and we include here the results for the Q-learning agent averaged over noise parameters $\epsilon = 0.2, 0.4, 0.6$. The mutual information per trajectory $m_i$ and cumulative mutual information $M_i$ are given in Figure~\ref{fig:gridworld-mutual}. Further experiments and experimental details can be found in the appendix, Section~\ref{sec:gridworld-appendix}.

\begin{figure}
  \centering
  \begin{subfigure}[m]{0.475\textwidth}
      \centering
      \includegraphics[width = \textwidth]{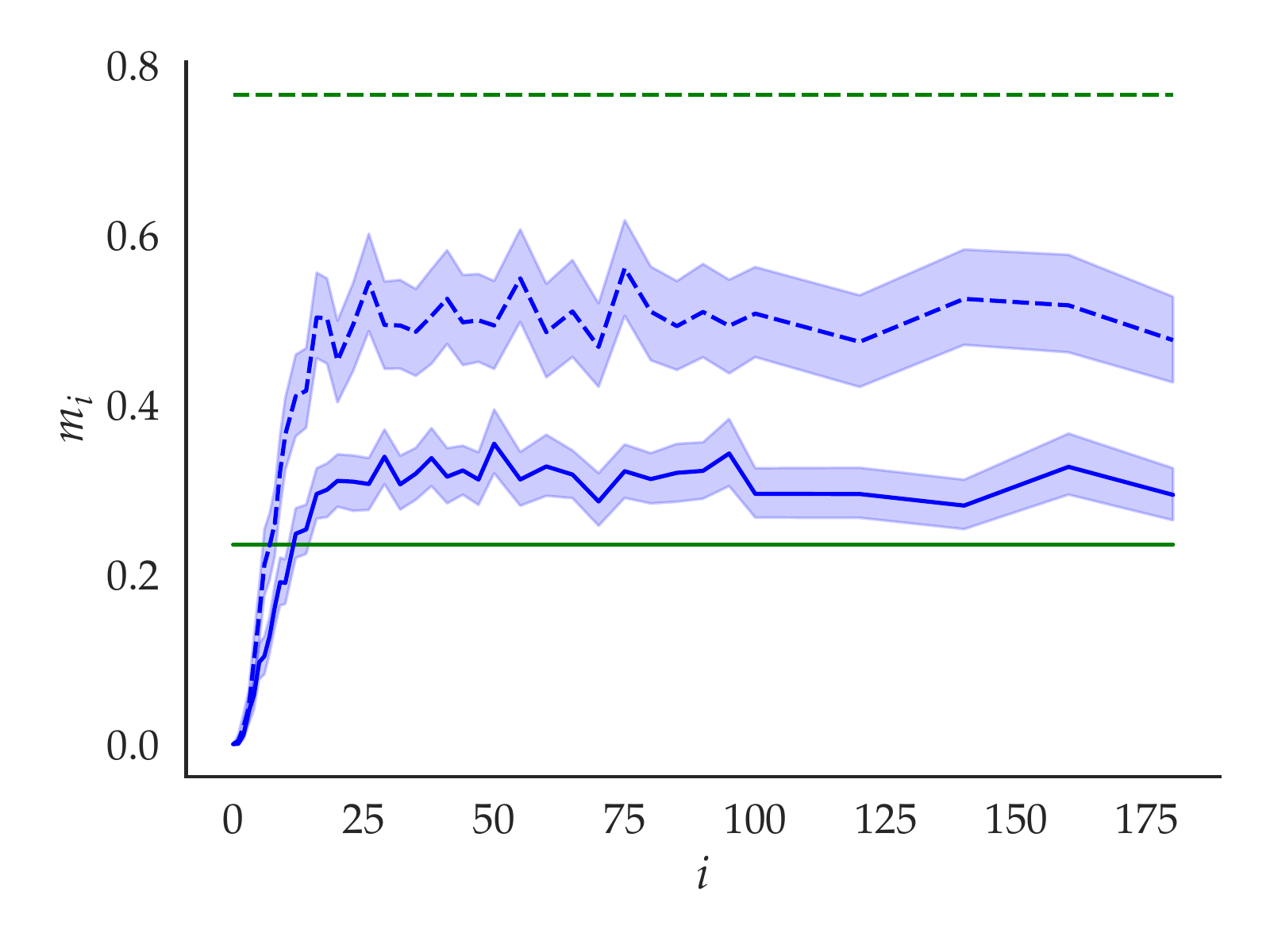}
  \end{subfigure}
  \begin{subfigure}[m]{0.475\textwidth}
      \centering
      \includegraphics[width=\textwidth]{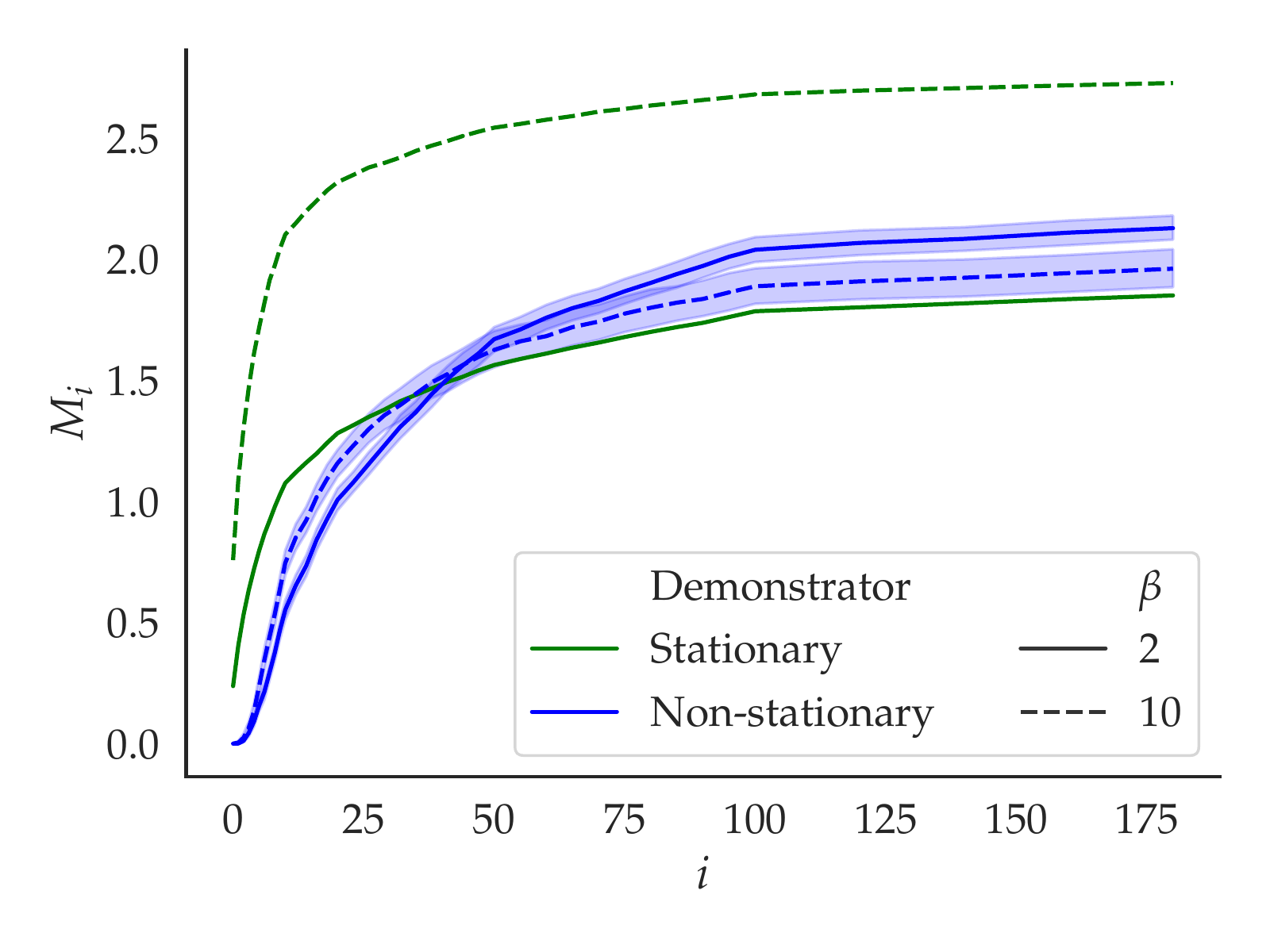}
  \end{subfigure}
  \caption{The per-timestep mutual information $m_i$ (left) and cumulative mutual information $M_i$ (right) between trajectory $\tau_i$ and reward parameter $\theta$, averaged over $\epsilon=0.2, 0.4, 0.6$, for stationary Boltzmann and non-stationary Q-learning \demonstrator{}, with 95\% bootstrapped confidence intervals. As discussed in Section~\ref{fig:gridworld}, we see that for low $\beta$ the non-stationary \demonstrator{} can be more informative than the Boltzmann \demonstrator{}.}\label{fig:gridworld-mutual}
\end{figure}

We see that for low $\beta = 2$ the non-stationary demonstrator can be more informative than the Boltzmann \demonstrator{}, whereas for high $\beta = 10$ the stationary \demonstrator{} outperforms. A lower $\beta$ leads to a weaker signal across the board as the \demonstrator{}'s actions exhibit more noise. In the appendix (Section~\ref{sec:gridworld-appendix}), we discuss how the non-stationary \demonstrator{} performance is robust to higher amounts of action noise $\epsilon$. We also discuss an observed phenomenon where mutual information seems to spike for earlier demonstrations in the case of the Direct Evaluation agent.

\subsubsection{Gridworld misspecification}\label{sec:misspec}

A natural question is whether modelling learning is necessary; perhaps a learning \demonstrator{} is just as informative when the observer is modelling them as stationary. To assess whether this is true, we look at the effect of having the wrong model for the Gridworld environment. When a model is misspecified, mutual information is not a valid measure of performance, instead we look at the posterior metrics $p_i, P_i$ introduced in Section~\ref{sec:metrics}.

\begin{wrapfigure}{r}{0.4\textwidth}
  \centering
  \includegraphics[width=0.35\textwidth]{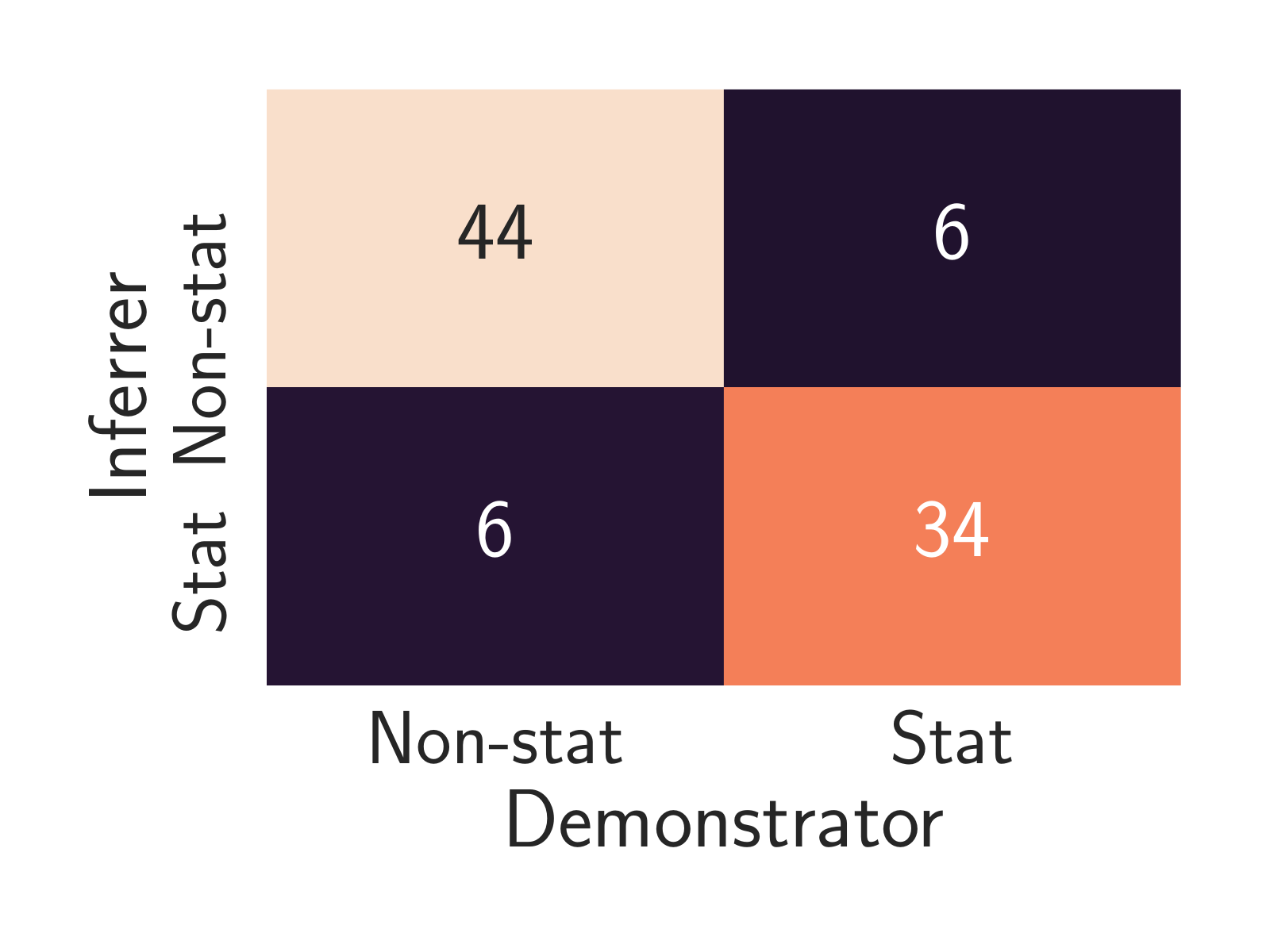}
  \caption{Posterior probabilities $P_{181}$ (\%) in the Gridworld domain for misspecified models among a stationary Boltzmann $D^\beta$ to Q-learning $L_q$, both with $\beta = 2$. We see that the off-diagonal does better than the prior of $4\%$, but barely so.}\label{fig:heatmap-small}
\end{wrapfigure}

First, we study the effects of making an incorrect assumption as to whether a \demonstrator{} is stationary or not. We run $n = 150$ experiments for each form of misspecification. The cumulative posterior after $181$ trajectories $P_{181}$ is given in Figure~\ref{fig:heatmap-small}. The per-trajectory posterior $p_i$ is given in Figure~\ref{fig:gridworld-misspec}. Further results and experimental details can be found in the appendix, Section~\ref{sec:gridworld-appendix}.

\begin{figure}
  \centering
  \includegraphics[width = \textwidth]{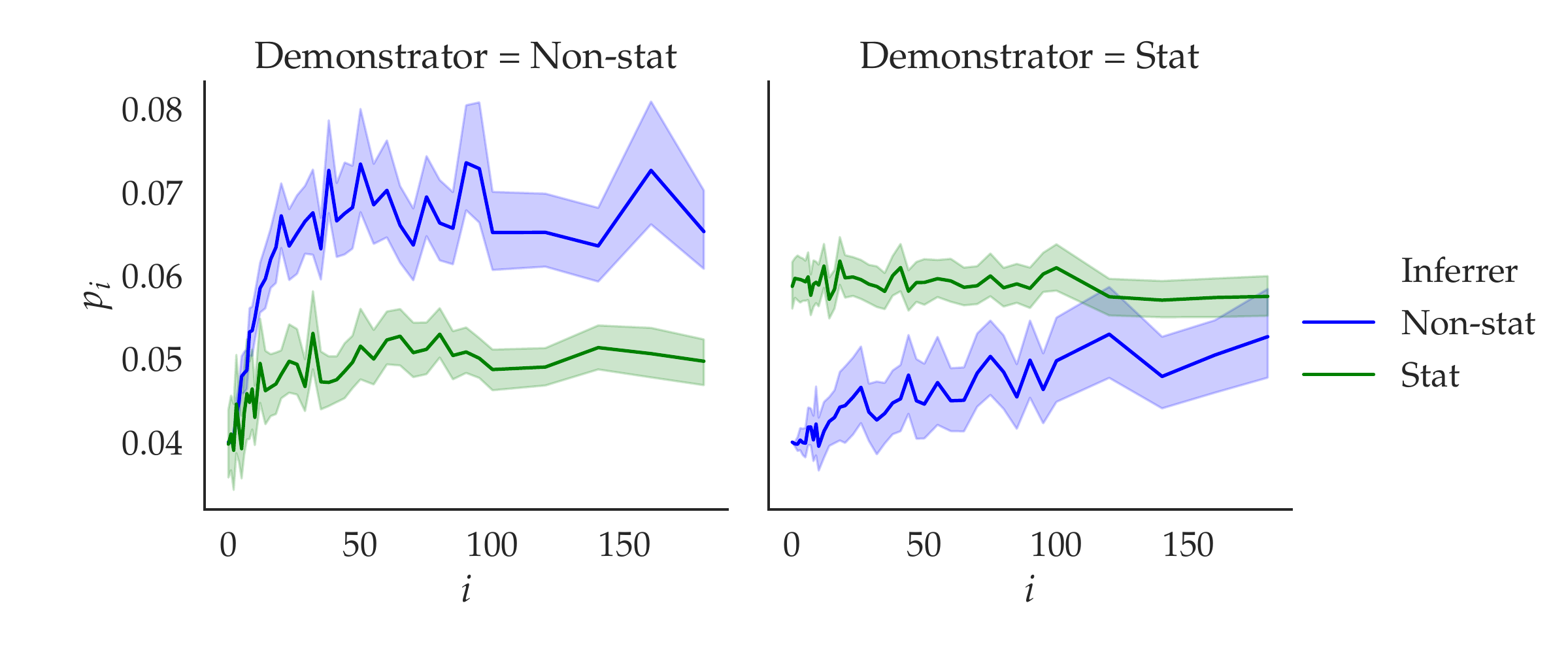}
  \caption{Per-trajectory posterior probabilities $p_i$ in the Gridworld domain for misspecified models among a stationary Boltzmann $D^\beta$ to Q-learning $L_q$, for $\beta = 2$, including a 95\% bootstrapped confidence interval. Results averaged over $\epsilon = 0.2, 0.4, 0.6$. As discussed in Section~\ref{sec:misspec}, we see that misspecification leads to poor inference, particularly in the early stages of demonstration.}\label{fig:gridworld-misspec}
\end{figure}

We see that misspecification leads to poor inference, only very marginally improving on the posterior of $4\%$. Furthermore, we see in Figure~\ref{fig:gridworld-misspec} that per-trajectory performance is worst for the earliest trajectories. In the case where the non-stationary \demonstrator{} is assumed to be stationary, inference improves over time as the agent converges to the policy expected by the \inferrer{}. This gives justification for the common practice in IRL demonstrations whereby the \demonstrator{} is given some time to understand the experiment setup and practice demonstrating before data collection begins (for example, see the instructions in \citet{christiano2017deep}). In the other case, where a stationary agent is assumed to be non-stationary, inference improves over time because the \inferrer{}, modelling the stationary \demonstrator{} as converging, is increasingly expecting the (near-)optimal behaviour that it is observing.

\begin{figure}
  \centering
  \includegraphics[width=0.6\textwidth]{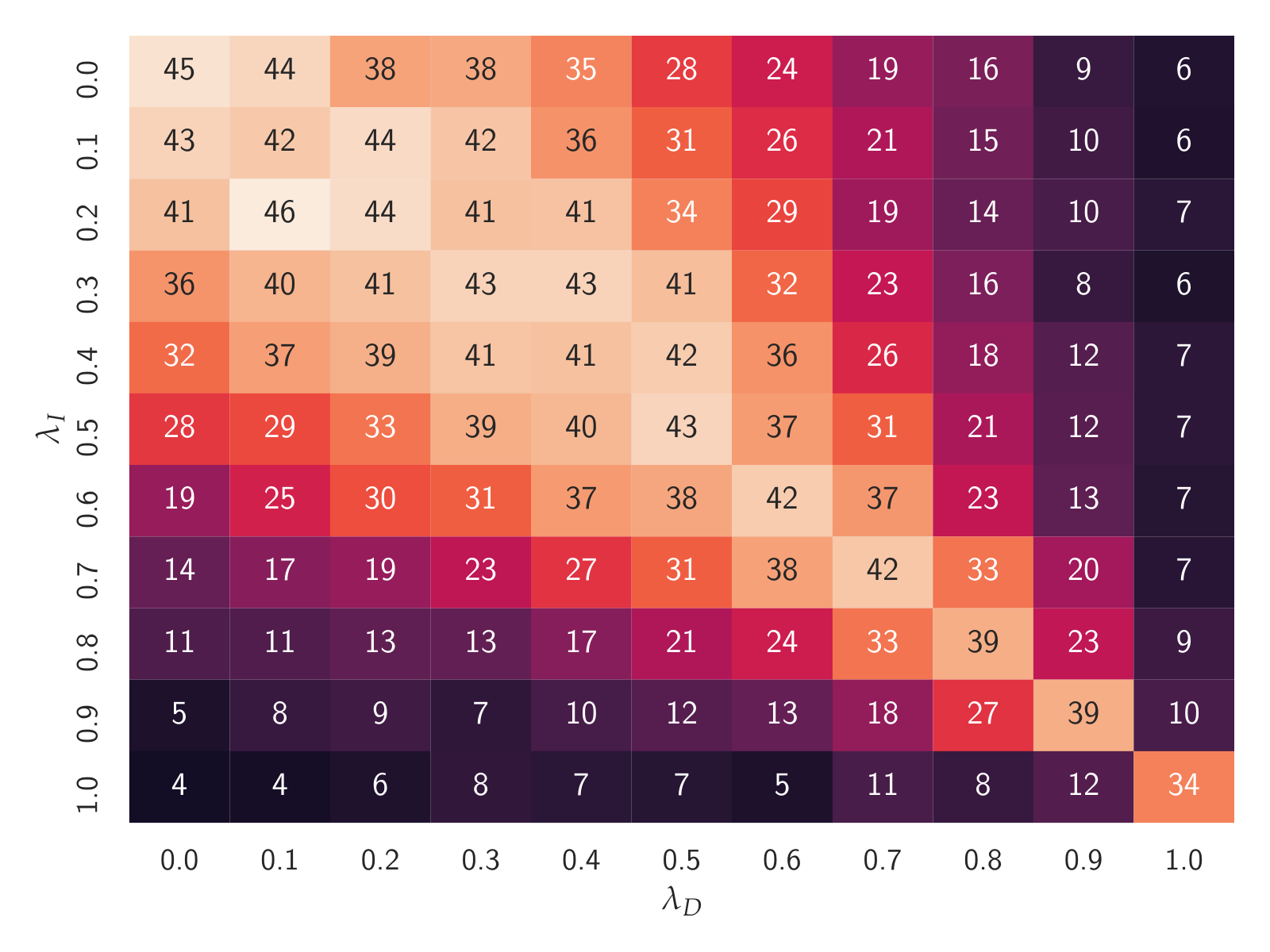}
  \caption{The posterior probability $P_{181}$ ($\%$) after 181 observed trajectories in the Gridworld domain. We observe that inference is robust to misspecificaiton of $\lambda$. For example, when the \demonstrator{} has parameter $\lambda_D = 0.5$, the \inferrer{} assigns a posterior $\geq 20\%$ to the true reward for $\lambda_I \in [0, 0.8]$.}\label{fig:heatmap}
\end{figure}

Second we look at the effects of correctly allowing for non-stationarity but having the incorrect model. We assess this by considering the TD$(\lambda)$ model $L_\lambda$ where the \demonstrator{} has parameter $\lambda = \lambda_D$ and the \inferrer{} has parameter $\lambda = \lambda_I$. We run $n = 150$ experiments for a mesh of values $\lambda_D, \lambda_I \in [0, 1]$. Figure~\ref{fig:heatmap} gives a heatmap of the cumulative posterior after a demonstration of $181$ trajectories, $P_{181}$.

We observe that inference is mostly robust to misspecification of $\lambda$. For example, when the \demonstrator{} has parameter $\lambda_D = 0.5$, the \inferrer{} assigns a posterior $\geq 20\%$ to the true reward for all $\lambda_I$ in the range $[0, 0.8]$. Only at the bottom left and bottom right (where the \inferrer{}'s model is most misspecified) is performance is comparable to the performance of mismodelling the \demonstrator{} as stationary. In each case, the \inferrer{} does better than the prior $4\%$.

\subsection{Random MDPs}\label{sec:random}
\begin{figure}
    \centering
    \begin{subfigure}[m]{0.43\textwidth}
        \centering
        \includegraphics[width = \textwidth]{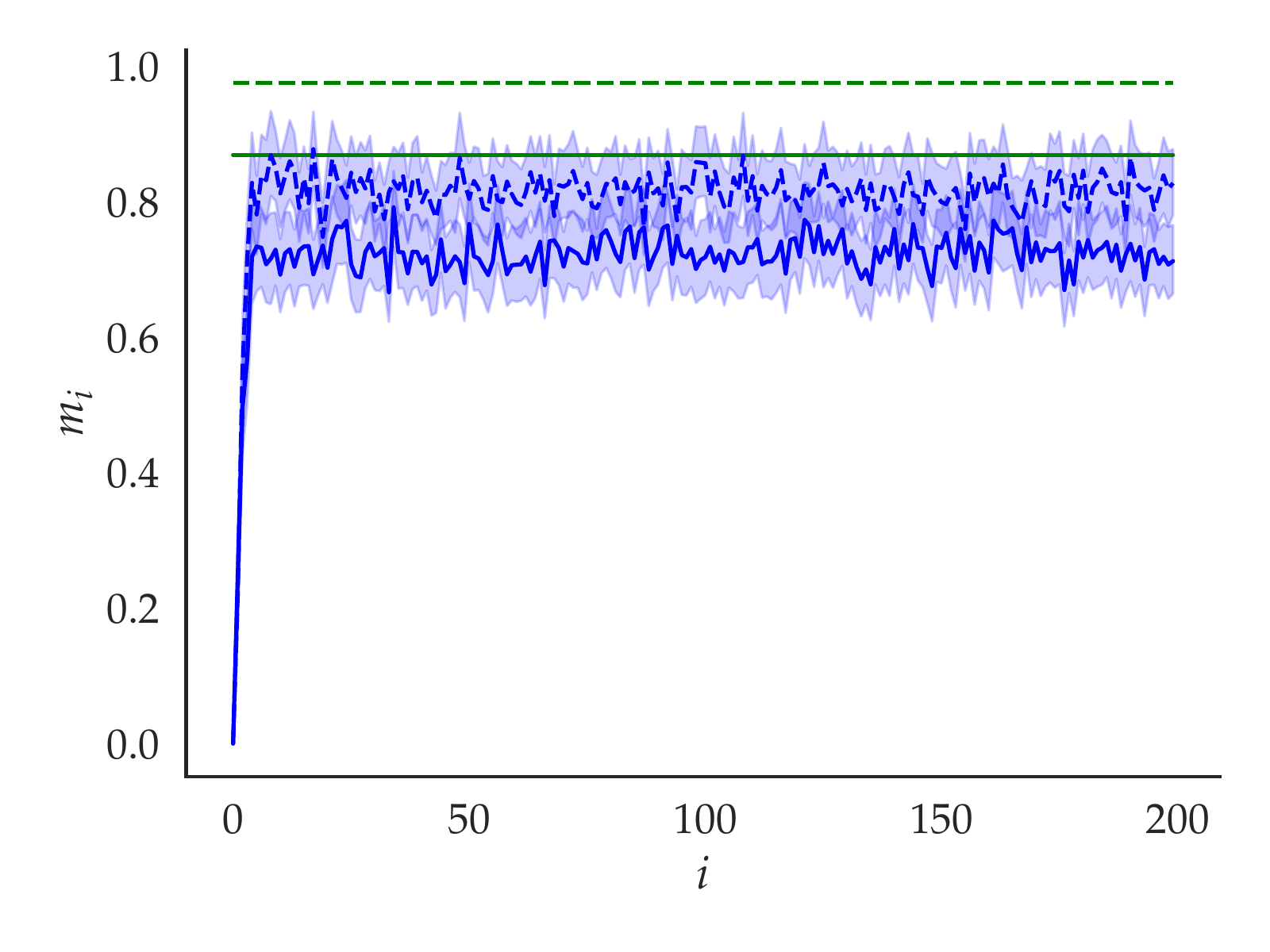}
    \end{subfigure}
    \begin{subfigure}[m]{0.43\textwidth}
        \centering
        \includegraphics[width=\textwidth]{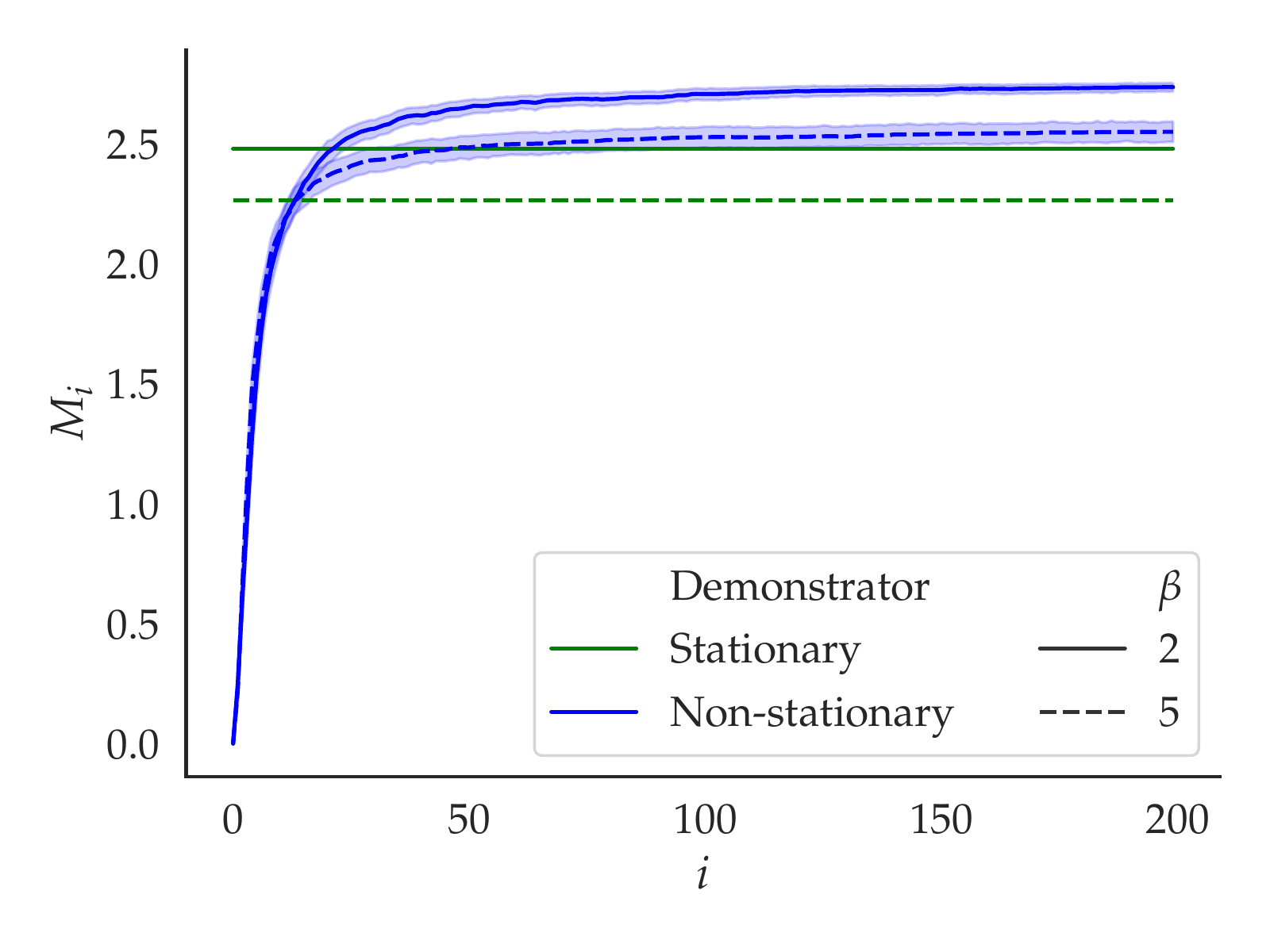}
    \end{subfigure}
    \caption{Mutual information for Random MDP experiments with $|\mathcal{S}| = 3$, $\beta = 2, 5$, comparing stationary Boltzmann \demonstrator{} with non-stationary Q-learning \demonstrator{}. Left: per-trjactory mutual information $m_i$ (higher is better) including 95\% bootstrapped confidence intervals. Right: cumulative mutual information $M_i$. As discussed in Section~\ref{sec:random}, we see that although per-trajectory information is lower for agents which are learning, the cumulative mutual information can be higher.}\label{fig:random-mdp-mutual}
\end{figure}

To explore a more general domain, next we consider randomly generated MDPs with $|\mathcal{A}| = 2$ and for $|\mathcal{S}| \in \{3, 4, 5, 6\}$. Two of the states generate reward in $\{-1, 1\}$, giving $|\Theta| = 16$. We run $n = 200$ experiments, each with a demonstration of 200 trajectories. The mutual information is given in Figure~\ref{fig:random-mdp-mutual} for $|\mathcal{S}| = 3$, and additional results are available in Section~\ref{sec:gridworld-appendix} of the appendix.

Interestingly, we find in all of the Random MDP experiments that although per-trajectory information is lower for the non-stationary agent, the more important measure of cumulative mutual information is higher. This highlights a crucial difference between the two models. The stationary agent is giving answers to a single highly informative question, namely a request for a (near-)optimal demonstration. Whereas the non-stationary \demonstrator{} is giving answers to a multitude of different --- in this case less informative --- questions, namely the \demonstrator{}'s preference for the trajectories experienced so far, as represented through changes in behaviour. 

\section{Limitations and Future work}
\textbf{Experiments with real humans.} As a first step toward studying the interaction of demonstrator learning with the reward learning of an inferrer, we conducted experiments entirely in simulation. Future work should perform experiments with real humans while finding ways to surmount the difficulties noted in Section~\ref{sec:why-simulation}. Furthermore, it should be explored as to whether the model used in this paper is in fact a suitable model for human learning.

\noindent \textbf{Larger domains and approximate inference} As we performed exact inference in this work, we were restricted to domains with a small set of discrete reward parameters. Future work could study how well our results hold up when we instead perform approximate inference on domains with larger reward parameter spaces, or model the learning algorithm also as a random variable among a family of algorithms.

\newpage
\section*{Broader Impact}
Many real-world tasks we would like AIs to do are difficult because they lack an easily-specified reward function. For example, it's very difficult to specify a reward function that would induce courteous behavior in a self-driving car, and even harder to make one that correctly balances all the trade-offs that a car would face in the real world~\cite{awad2018moral}. Reward inference is one approach to tackling this problem --- instead of requiring a designer to specify the reward, it only requires a designer to specify a prior over reward functions and provide a dataset from which to learn the reward.

Thus, any work that contributes to better reward inference contributes to building better AIs in these real-world tasks, and our work is no exception. As an example of the possible applications of our work, better understanding what the users of our AI desire could lead to recommender systems that can optimize for individual user preferences and autonomous cars that adapt to the risk and time preferences of their riders.

However, better reward inference could be used by bad actors in malicious ways. For example, many intentionally malicious behaviors may be hard to specify manually. Just like it is hard to specify the reward for a courteous car, it is hard to specify the reward of a purposely rude car. Better reward inference would make it easier to specify either type of behavior.

A better understanding of what individual humans want could also be used to manipulate their behavior. For example, a better model of person's preferences for internet browsing could be used to develop more effective personalized online scams or spear phishing attacks.


\bibliography{main.bib}
\bibliographystyle{plainnat}

\newpage
\appendix
\section{Gridworld experiments}\label{sec:gridworld-appendix}

Our Gridworld experiments were run for $\epsilon \in \{0.2, 0.3, 0.4, 0.45, 0.49, 0.6, 0.65, 0.66\}$, for stationary Boltzmann \demonstrator{} $D^\beta$ and learning algorithms $L_q, L_d$. The initial state distribution is uniform over the non-terminal states. We use $\gamma = 0.98$. Trajectories were cutoff at a maximum length of 100 timesteps. Value iteration for stationary agents was run till convergence, with a tolerance of $10^{-9}$ for maximum Q-value divergence between iteration steps.

\begin{figure}
  \centering
  \includegraphics[width = \textwidth]{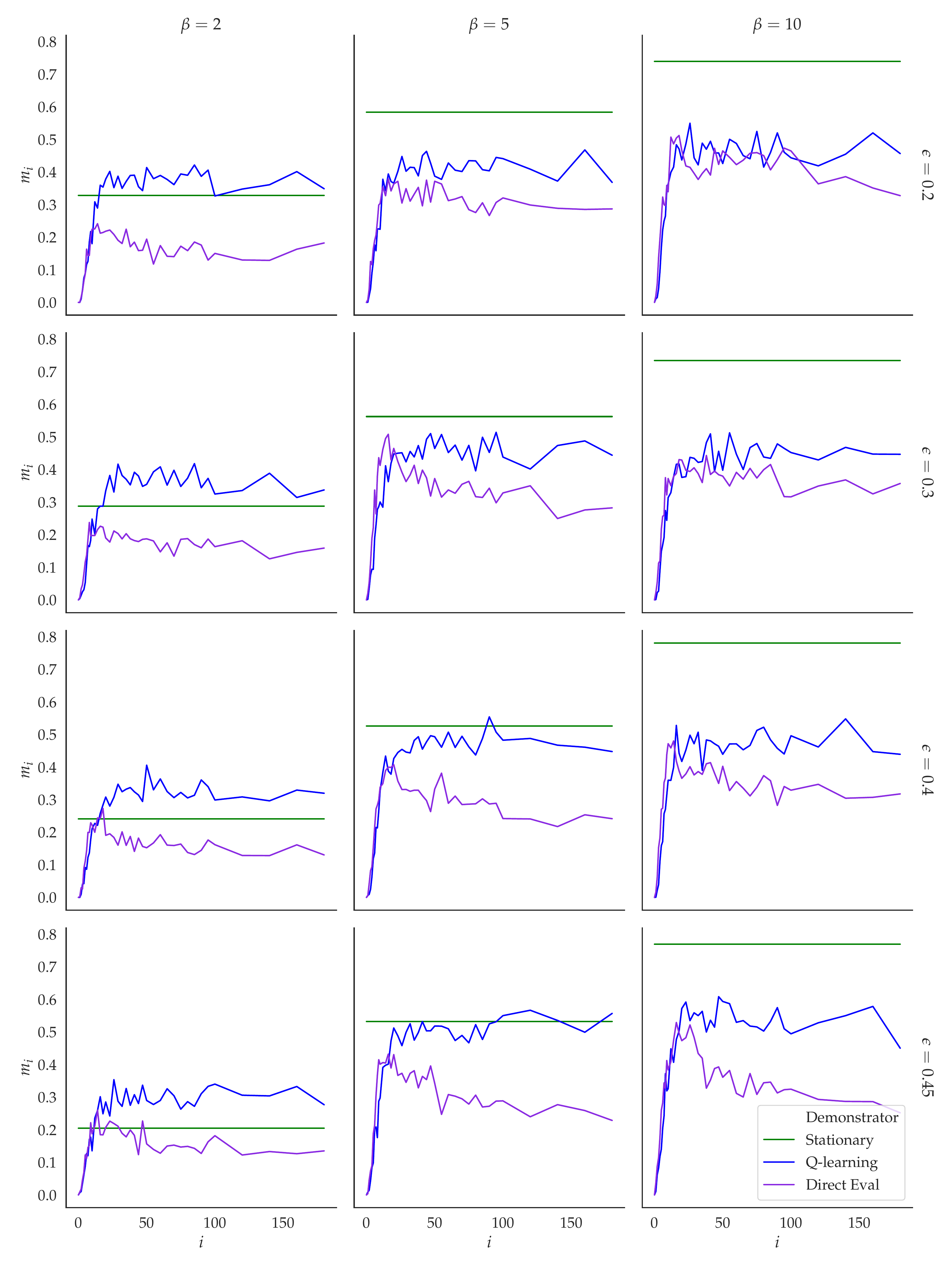}
  \caption{Mutual information per timestep $m_i$ among a variety of $\epsilon$ and $\beta$, for specified experiments with the \demonstrator{} being either stationary Boltzmann, Q-learning, or Direct Evaluation.}\label{fig:grid-appendix-0}
\end{figure}

\begin{figure}
  \centering
  \includegraphics[width = \textwidth]{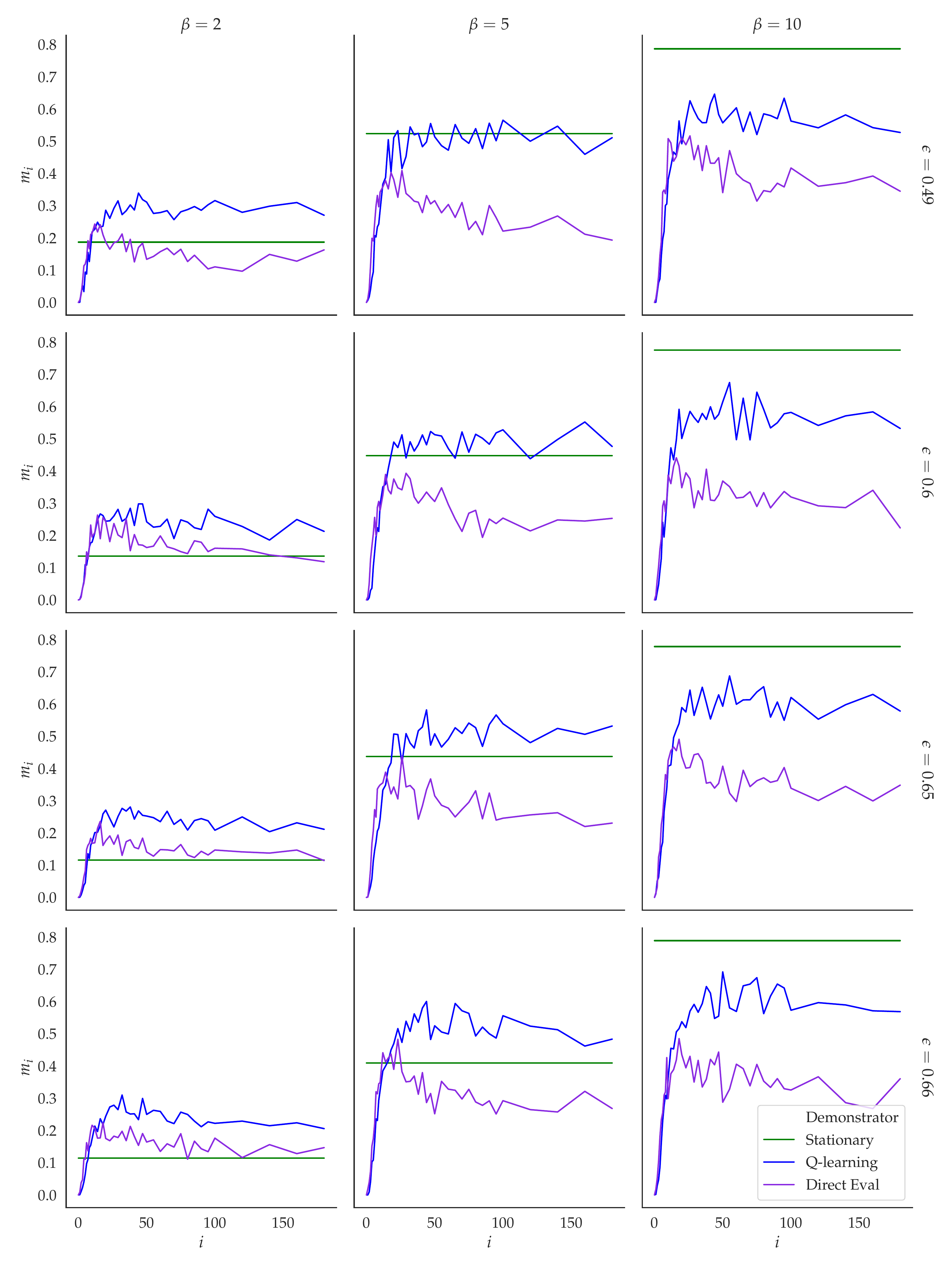}
  \caption{Mutual information per timestep $m_i$ among a variety of $\epsilon$ and $\beta$, for specified experiments with the \demonstrator{} being either stationary Boltzmann, Q-learning, or Direct Evaluation.}\label{fig:grid-appendix-1}
\end{figure}

\begin{figure}
  \centering
  \includegraphics[width = \textwidth]{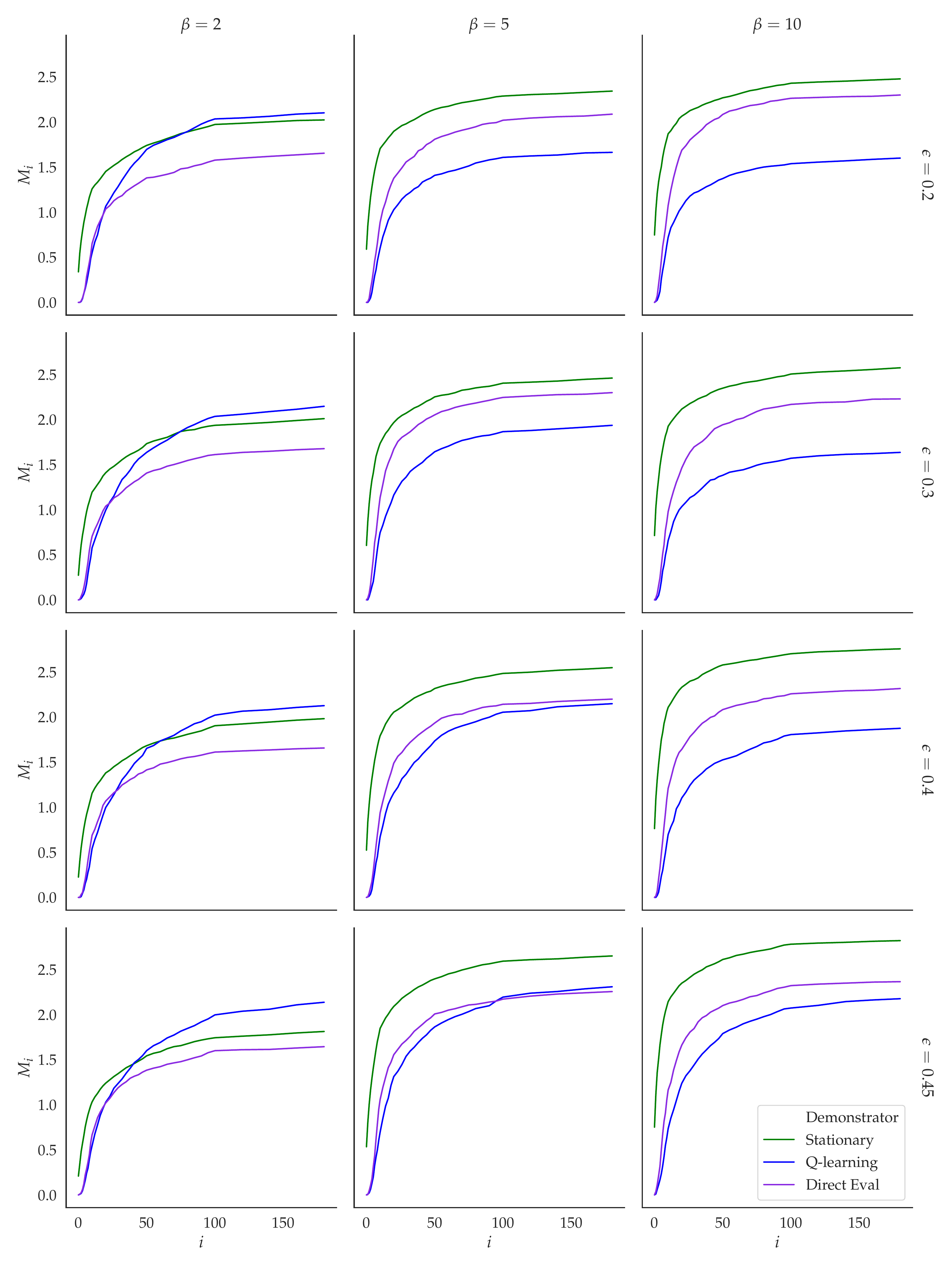}
  \caption{Cumulative mutual information $M_i$ among a variety of $\epsilon$ and $\beta$, for specified experiments with the \demonstrator{} being either stationary Boltzmann, Q-learning, or Direct Evaluation.}\label{fig:grid-appendix-2}
\end{figure}

\begin{figure}
  \centering
  \includegraphics[width = \textwidth]{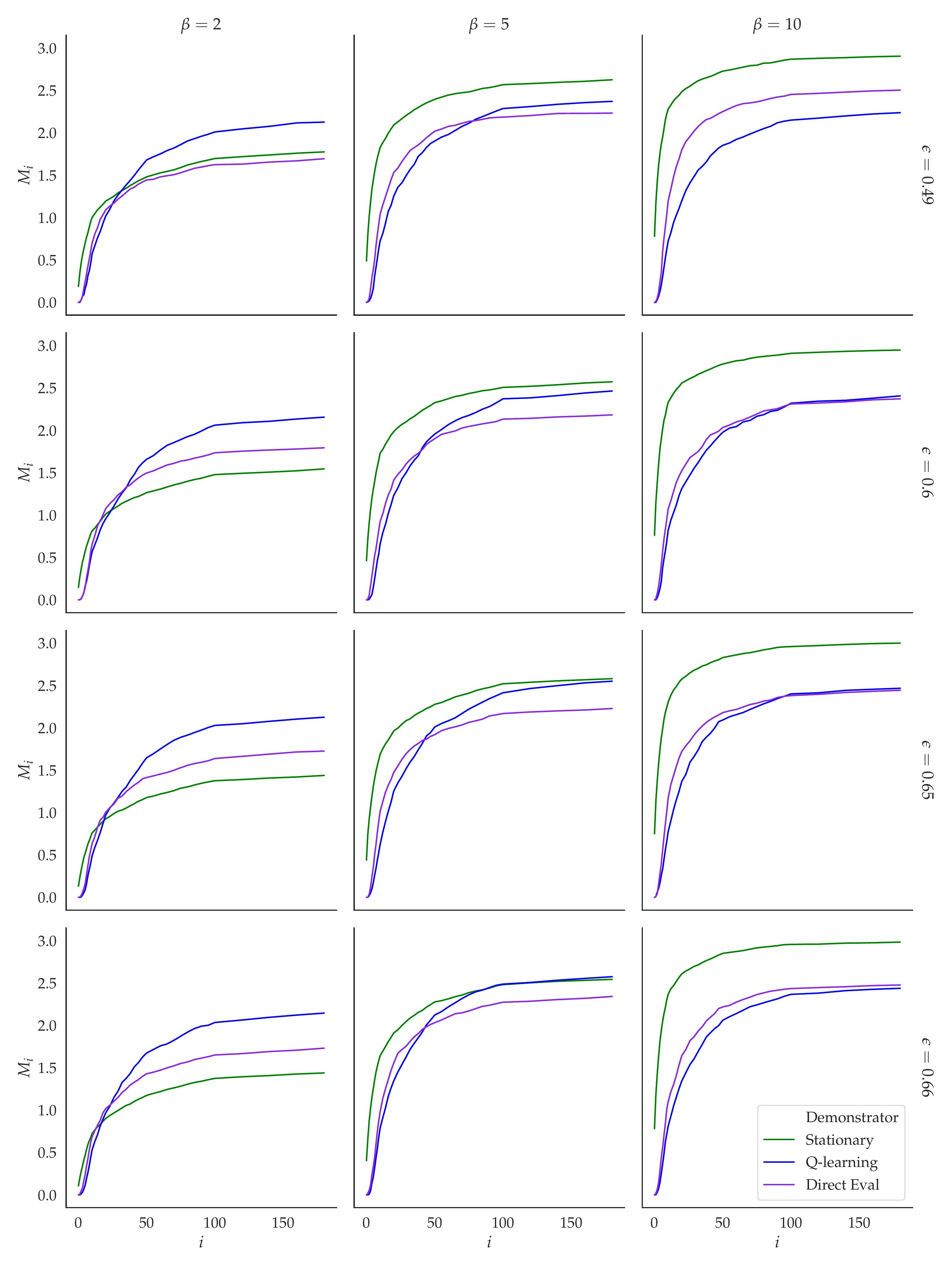}
  \caption{Cumulative mutual information $M_i$ among a variety of $\epsilon$ and $\beta$, for specified experiments with the \demonstrator{} being either stationary Boltzmann, Q-learning, or Direct Evaluation.}\label{fig:grid-appendix-3}
\end{figure}

\begin{figure}
  \centering
  \includegraphics[width = \textwidth]{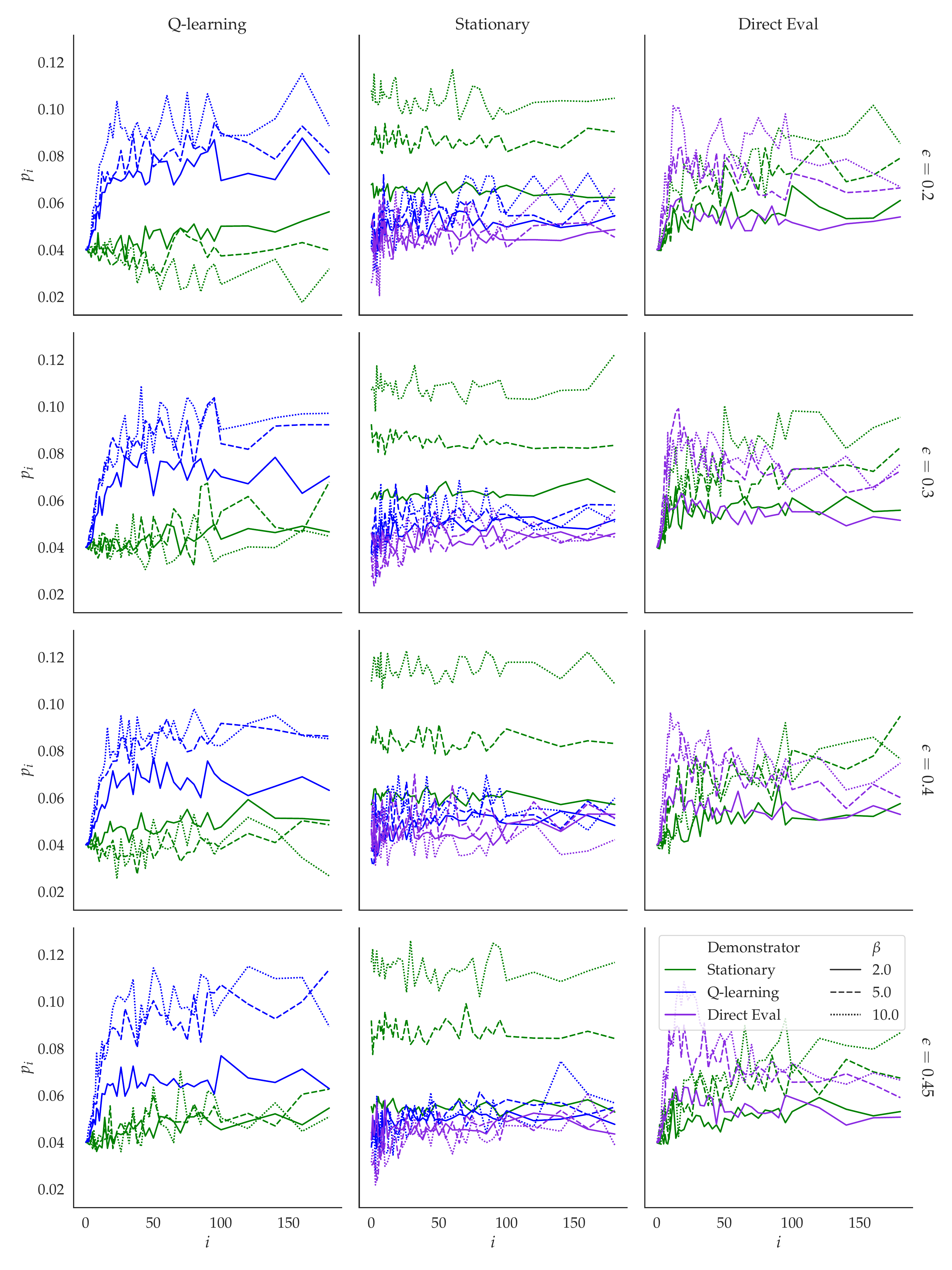}
  \caption{Posterior per timestep $p_i$ among a variety of $\epsilon$ and $\beta$, for misspecified experiments with the \demonstrator{} being either stationary Boltzmann, Q-learning, or Direct Evaluation.}\label{fig:grid-appendix-4}
\end{figure}

\begin{figure}
  \centering
  \includegraphics[width = \textwidth]{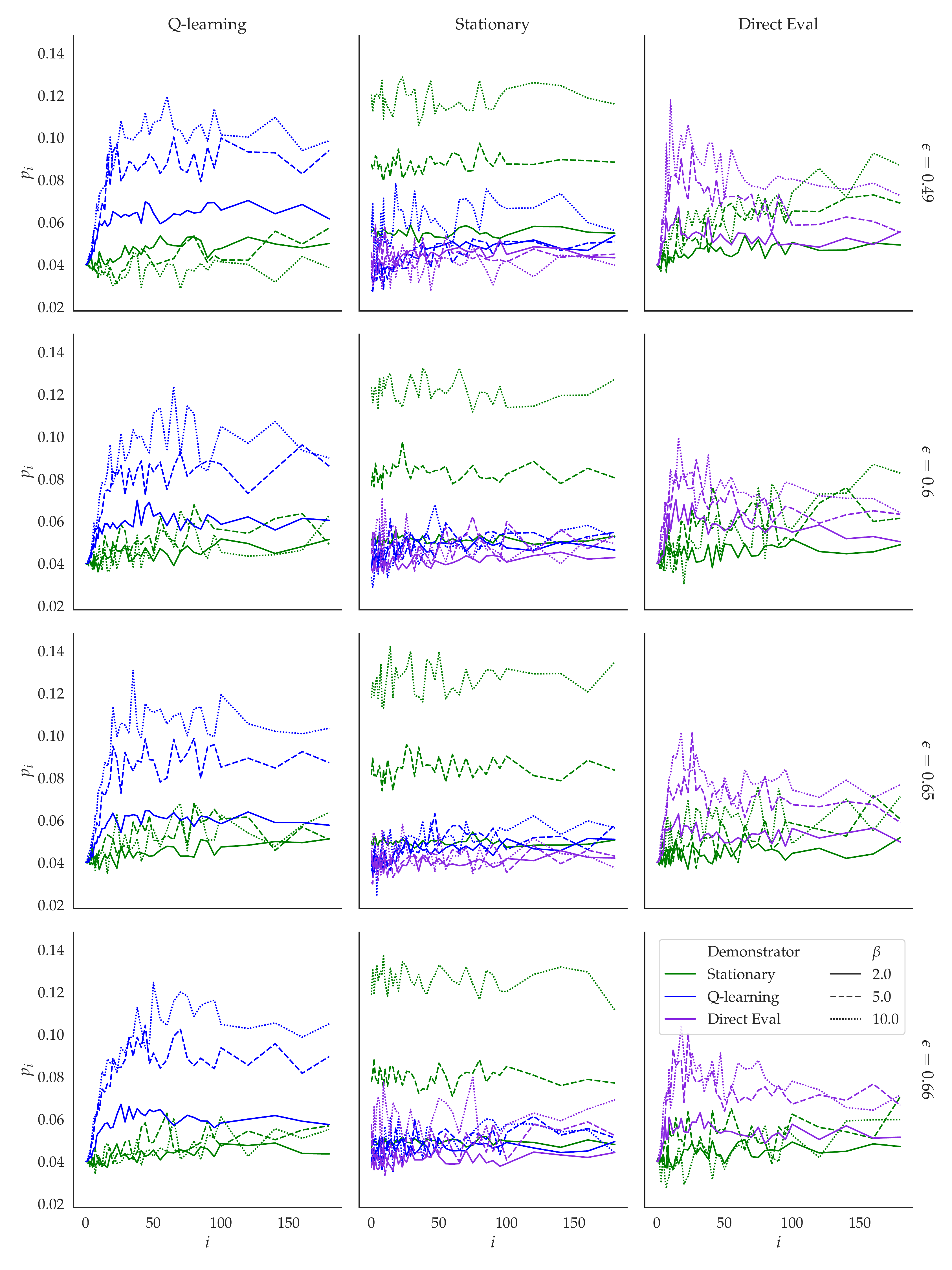}
  \caption{Posterior per timestep $p_i$ among a variety of $\epsilon$ and $\beta$, for misspecified experiments with the \demonstrator{} being either stationary Boltzmann, Q-learning, or Direct Evaluation.}\label{fig:grid-appendix-5}
\end{figure}

In the main body we include results for $L_q$, and averaged over $\epsilon = 0.2, 0.4, 0.6$. We note that mutual information is generally higher for $L_q$ compared with $L_d$. To see a full array of results consult Figures~\ref{fig:grid-appendix-0},~\ref{fig:grid-appendix-1},~\ref{fig:grid-appendix-2},~\ref{fig:grid-appendix-3}. For misspecification results consult Figures~\ref{fig:grid-appendix-4},~\ref{fig:grid-appendix-5},~\ref{fig:grid-appendix-6},~\ref{fig:grid-appendix-7}.

We see from these results that as with the illustrative examples, more noise leads to lower mutual information in the stationary case. The information from a Direct Evaluation \demonstrator{} does not vary much with $\epsilon$, however the Q-learning \demonstrator{} appears to \textit{improve} as more noise is introduced. We suspect that this is in part because more noise leads to slower convergence and more changes in behaviour. It is also evident, more-so from the misspecification plots Figures~\ref{fig:grid-appendix-6},~\ref{fig:grid-appendix-7}, that stationary inference is more sensitive to $\beta$ compared with both non-stationary models. In our experiments $\beta$ is always correctly specified, but this suggests that non-stationary might suffer less from a misspecified $\beta$.

We emphasise another feature of the Direct Evaluation agents where there is a spike in mutual information towards the beginning of the demonstration, see Figures~\ref{fig:grid-appendix-0},~\ref{fig:grid-appendix-1}. Attributing behaviour to a specific past experience is easier in the early stages when the list is smaller. Here this is overall more informative than later demonstrations for which individual trajectories have less impact on specific decisions.

\section{Random MDP experiments}\label{sec:random-appendix}

\begin{figure}
  \centering
  \includegraphics[width = \textwidth]{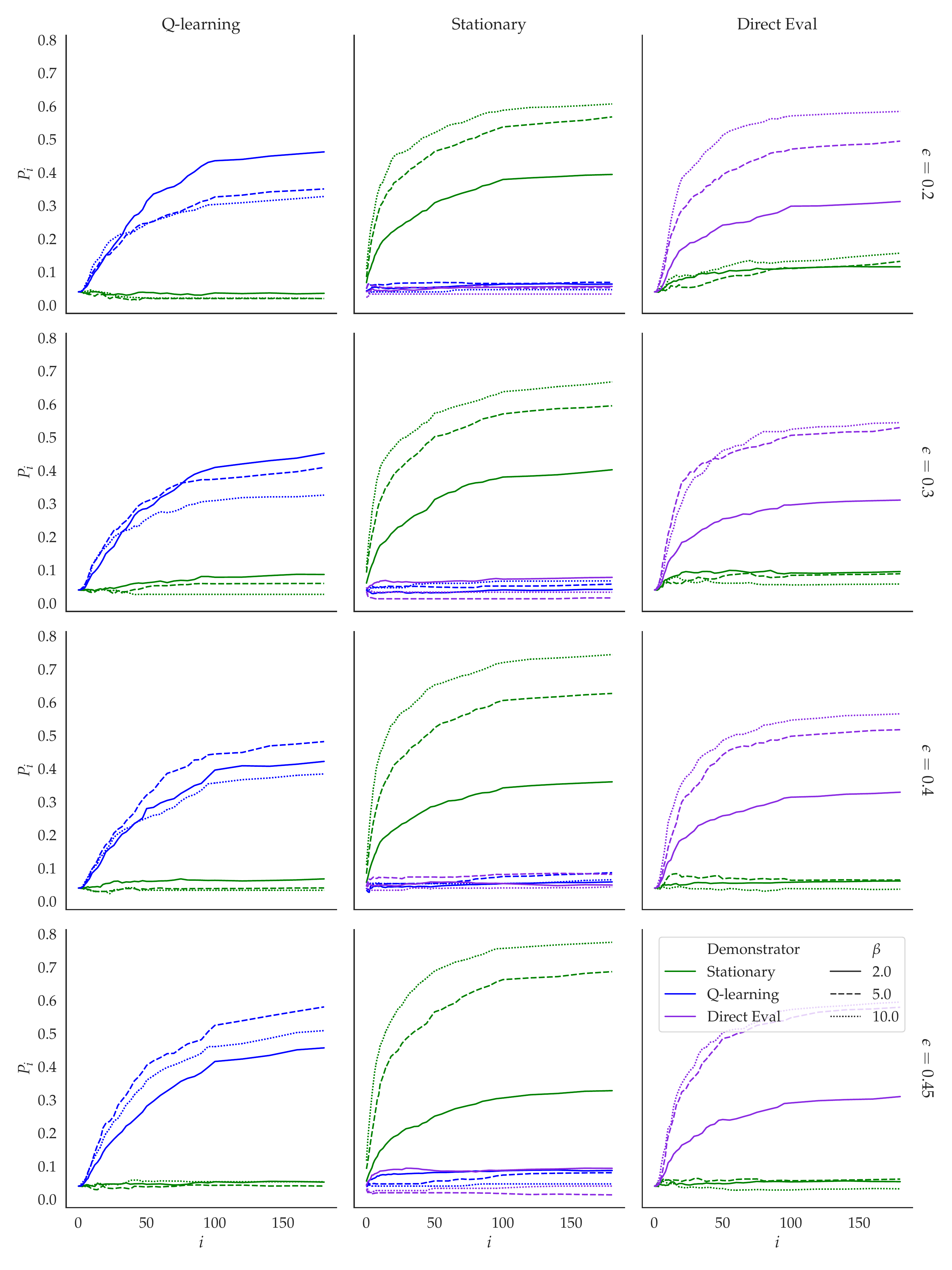}
  \caption{Cumulative posterior $P_i$ among a variety of $\epsilon$ and $\beta$, for misspecified experiments with the \demonstrator{} being either stationary Boltzmann, Q-learning, or Direct Evaluation.}\label{fig:grid-appendix-6}
\end{figure}

\begin{figure}
  \centering
  \includegraphics[width = \textwidth]{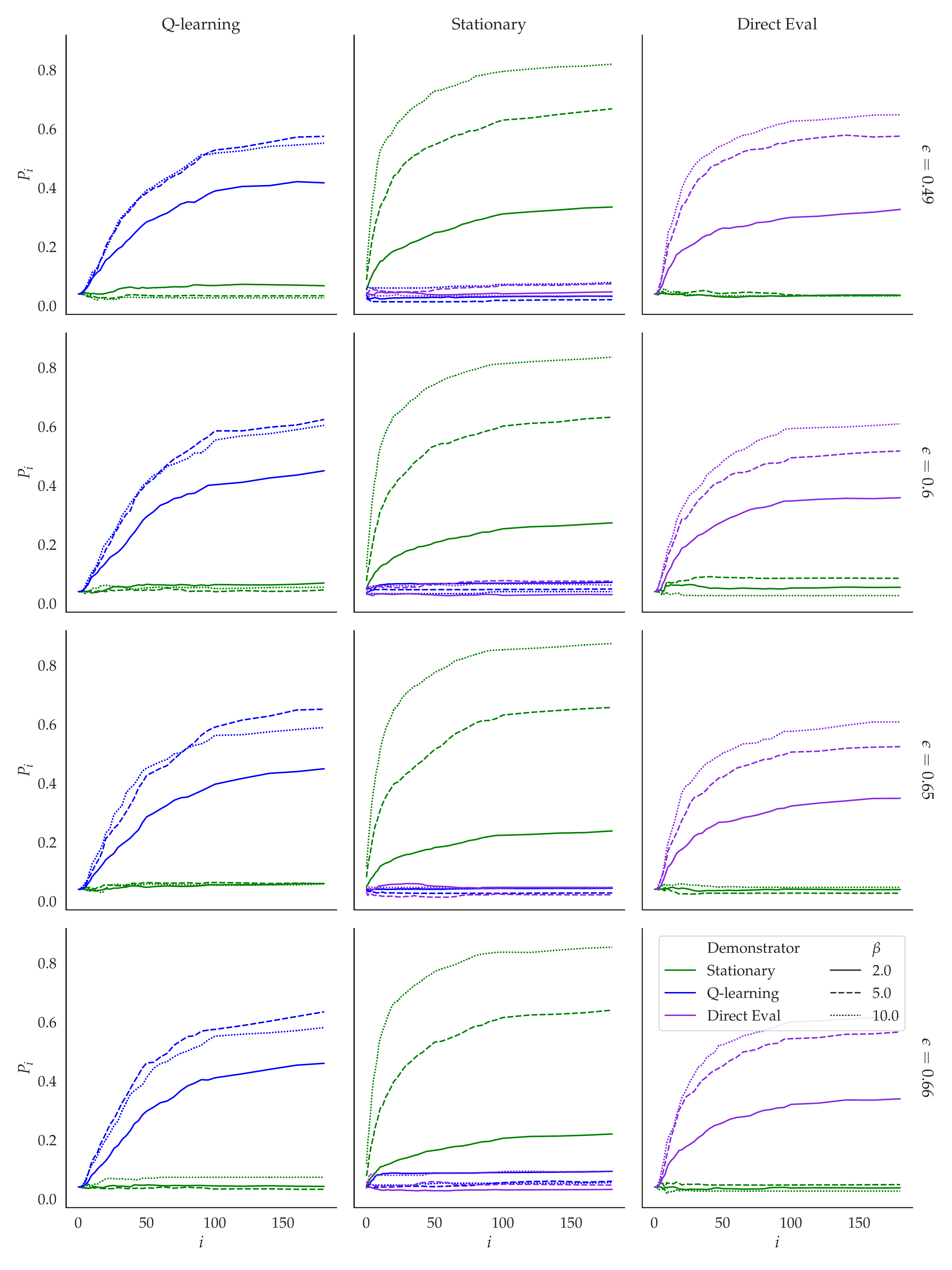}
  \caption{Cumulative posterior $P_i$ among a variety of $\epsilon$ and $\beta$, for misspecified experiments with the \demonstrator{} being either stationary Boltzmann, Q-learning, or Direct Evaluation.}\label{fig:grid-appendix-7}
\end{figure}

Our random MDP experiments were run for $|\mathcal{S}| \in \{3, 4, 5, 6\}$, for a stationary Boltzmann \demonstrator{} $D^\beta$ and learning algorithms $L_q, L_d$. The initial state distribution is uniform over the non-terminal states. We use $\gamma = 0.5$. Trajectories were cutoff at a maximum length of 100 timesteps. Value iteration for stationary agents was run till convergence, with a tolerance of $10^{-9}$ for maximum Q-value divergence between iteration steps. To see a full array of results consult Figures~\ref{fig:grid-appendix-6},~\ref{fig:grid-appendix-7}.

As with the Gridworld domain, we see a repeat of the phenomenon of the information spike for both the Direct Evaluation agent and the Q-learning agent, although it is less strong in the latter case. The effect appears to be stronger for higher $|\mathcal{S}|$, where the environment is more complex.

Mutual information in general appears to increase with $|\mathcal{S}|$. This is not surprising since with more starting states, and a larger set of possible trajectories, there are more ways for the \demonstrator{} to signal their preferences.

As discussed in Section~\ref{sec:random}, the Q-learning \demonstrator{} outperforms the stationary \demonstrator{} in each case. We see here that this cumulative outperformance of occurs very early on, in most cases within the first 30 trajectories.

As with the Gridworld domain, per-trajectory mutual information appears generally to be higher for higher Boltzmann parameter $\beta$, however in this case we see that cumulative mutual information is generally lower. Similar to the effect discussed in~\ref{sec:random}, this is because the \demonstrator{} is very effectively signalling a near the optimal trajectory, but the scope of this information is narrow and does not reflect enough the reward along suboptimal trajectories. In this domain, this results in less overall mutual information.

\begin{figure}
  \centering
  \includegraphics[width = \textwidth]{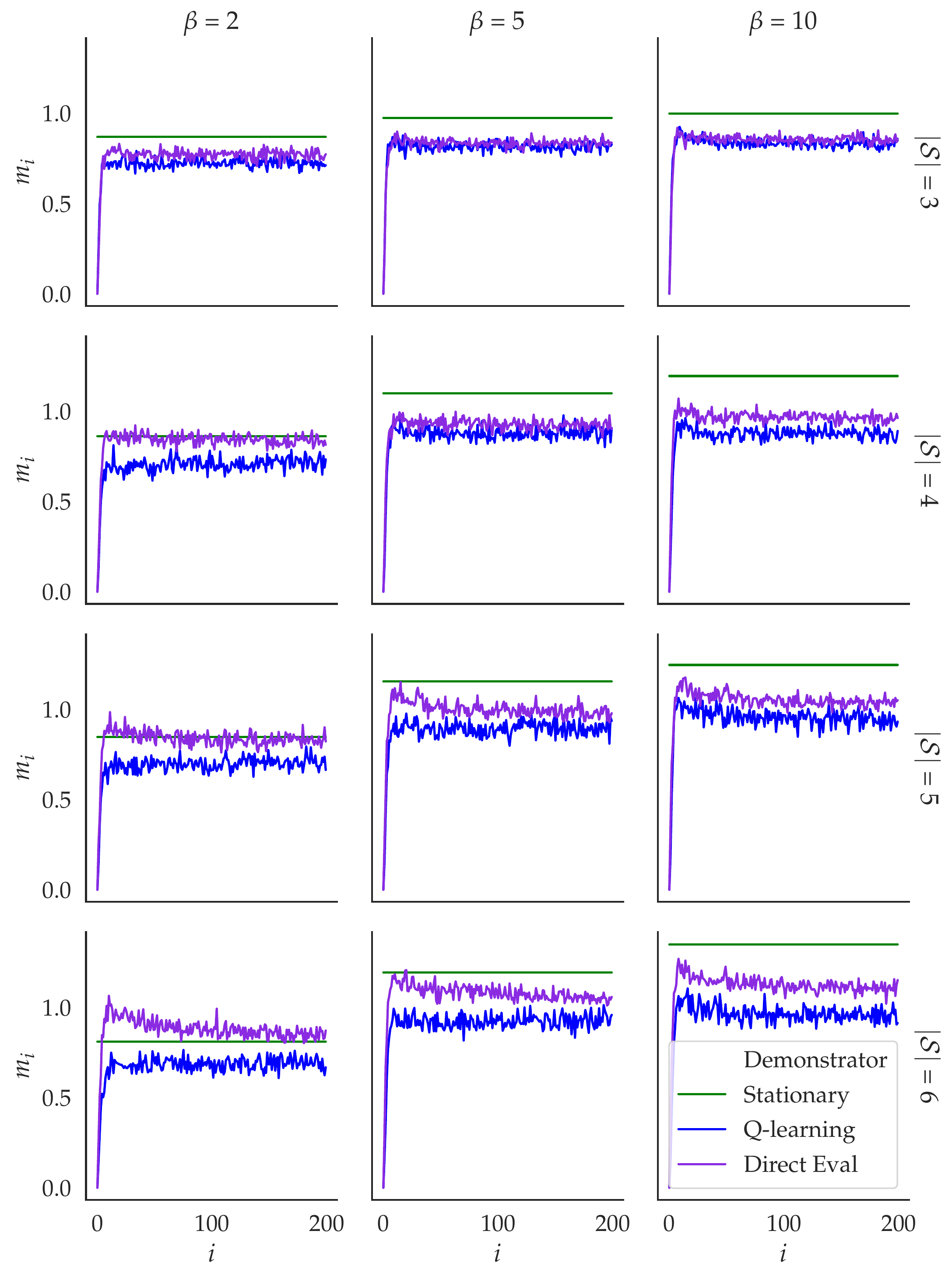}
  \caption{Per-trajectory mutual information $m_i$ among a variety of $|\mathcal{S}|$ and $\beta$ for the Random MDP domain with \demonstrator{} being either stationary Boltzmann, Q-learning, or Direct Evaluation.}\label{fig:random-appendix-0}
\end{figure}

\begin{figure}
  \centering
  \includegraphics[width = \textwidth]{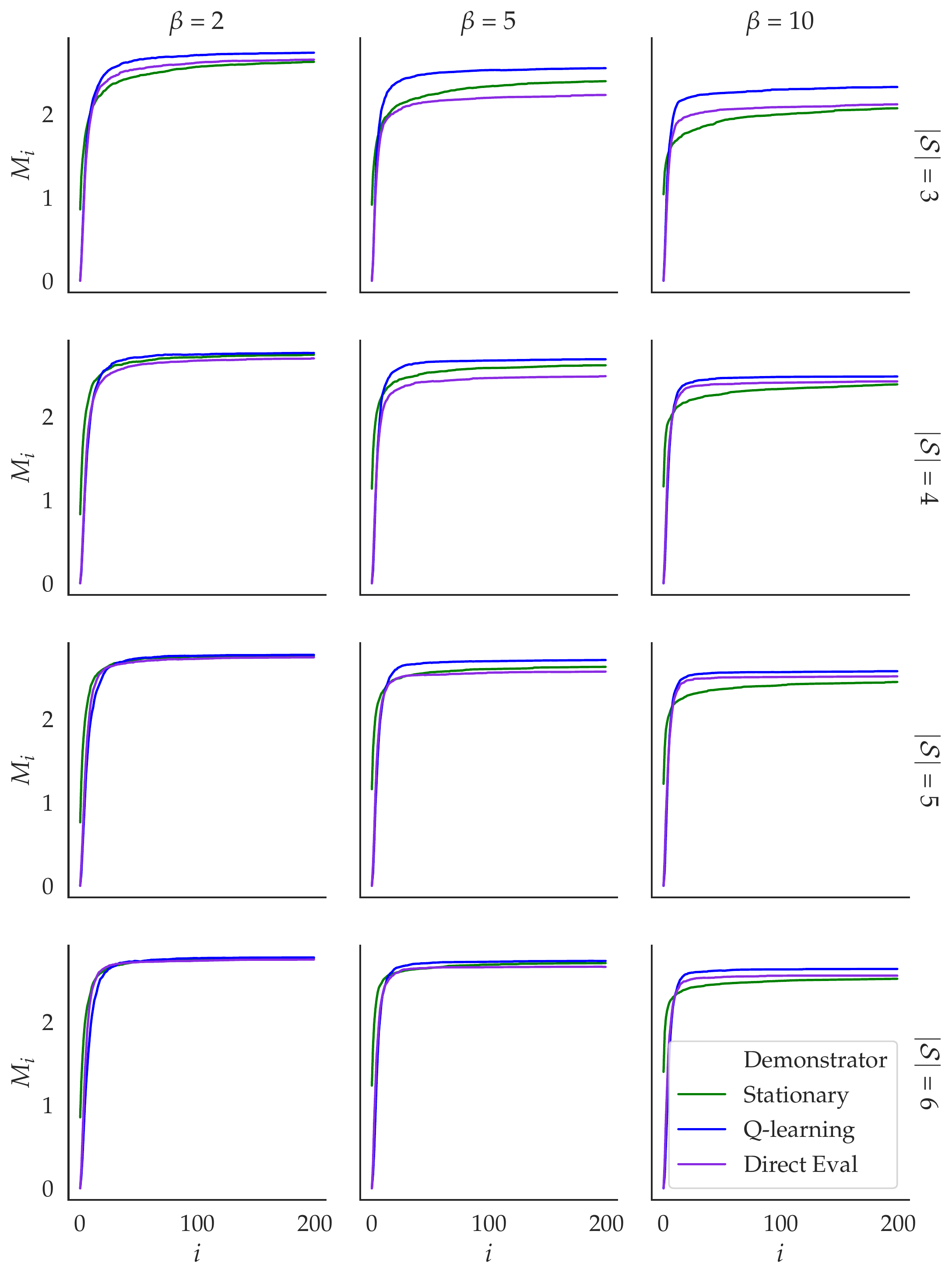}
  \caption{Cumulative mutual information $M_i$ among a variety of $|\mathcal{S}|$ and $\beta$ for the Random MDP domain with \demonstrator{} being either stationary Boltzmann, Q-learning, or Direct Evaluation.}\label{fig:random-appendix-1}
\end{figure}

\end{document}